\documentclass{article}

\usepackage{PRIMEarxiv}

\usepackage[utf8]{inputenc} % allow utf-8 input
\usepackage[T1]{fontenc}    % use 8-bit T1 fonts
\usepackage{hyperref}       % hyperlinks
\usepackage{url}            % simple URL typesetting
\usepackage{booktabs}       % professional-quality tables
\usepackage{amsfonts}       % blackboard math symbols
\usepackage{nicefrac}       % compact symbols for 1/2, etc.
\usepackage{microtype}      % microtypography
\usepackage{lipsum}
\usepackage{fancyhdr}       % header
\usepackage{graphicx}       % graphics
\graphicspath{{media/}}     % organize your images and other figures under media/ folder

\usepackage{amsmath,amssymb,amsfonts}
\usepackage{soul}
\usepackage{tabularx}
\usepackage{subfigure}
\usepackage{color}
\usepackage{multirow}

\newcommand{\etal}{\textit{et al}. }
\newcommand{\ie}{\textit{i}.\textit{e}., }

\title{Flexible Style Image Super-Resolution using Conditional Objective
}

\author{
  Seung Ho Park$^{1,2}$, Young Su Moon$^2$, Nam Ik Cho$^1$ \\
  $^{1}$Department of Electrical and Computer Engineering, INMC\\
Seoul National University, Seoul, 08826 Korea.\\
  $^{2}$Visual Display Division, Samsung Electronics, Suwon-si, 16677 Korea.\\
  \texttt{seungho@snu.ac.kr, mys66@samsung.com, nicho@snu.ac.kr}\\
}
\begin{document}
\maketitle

\begin{abstract}
Recent studies have significantly enhanced the performance of single-image super-resolution (SR) using convolutional neural networks (CNNs). While there can be many high-resolution (HR) solutions for a given input, most existing CNN-based methods do not explore alternative solutions during the inference. A typical approach to obtaining alternative SR results is to train multiple SR models with different loss weightings and exploit the combination of these models. Instead of using multiple models, we present a more efficient method to train a single adjustable SR model on various combinations of losses by taking advantage of multi-task learning. Specifically, we optimize an SR model with a conditional objective during training, where the objective is a weighted sum of multiple perceptual losses at different feature levels. The weights vary according to given conditions, and the set of weights is defined as a style controller. Also, we present an architecture appropriate for this training scheme, which is the Residual-in-Residual Dense Block equipped with spatial feature transformation layers. At the inference phase, our trained model can generate locally different outputs conditioned on the style control map. Extensive experiments show that the proposed SR model produces various desirable reconstructions without artifacts and yields comparable quantitative performance to state-of-the-art SR methods. Code and trained models will be available at \href{https://github.com/seungho-snu/FxSR}{https://github.com/seungho-snu/FxSR}
\end{abstract}

% keywords can be removed

%=====================
\section{Introduction}
Finding a high-resolution (HR) counterpart from a given low-resolution (LR) image is referred to as single image super-resolution (SISR). The SISR is an ill-posed problem in that infinitely many HR images correspond to a single LR image. Despite such ill-posedness, recent convolutional neural networks (CNNs) are shown to map an LR to a plausible HR~\cite{yang2019deep}.
%------------------
SRCNN~\cite{dong2014learning, dong2015image} first showed the effectiveness of a CNN for SISR, and various CNN architectures have been proposed for better performance afterward~\cite{kim2016accurate, kim2016deeply, shi2016real, tai2017memnet, lim2017enhanced, zhang2018image, zhang2018residual, yang2018drfn, jin2019flexible, zhang2019single, he2019mrfn, zhang2020accurate, tian2020coarse}. Earlier works used mean square error (MSE) as a loss function to train the network. However, since it tends to produce blurry HR outputs, researchers are finding new loss functions to generate more realistic outputs~\cite{928e476715544027af08ec20936dd6ca, dosovitskiy2016generating}. Specifically, perceptual losses ~\cite{johnson2016perceptual} are introduced to optimize the super-resolution (SR) model in the feature space instead of pixel space. Ledig \etal~\cite{2017photo} proposed to use adversarial loss~\cite{goodfellow2014generative} in combination with the perceptual loss to encourage the network to favor perceptually superior solutions residing in the manifold of natural images.
%------------------
More recently, Wang \etal~\cite{2018recovering} investigated class-conditional SR. It employed Spatial Feature Transform (SFT) capable of altering an SR network's behavior conditioned on semantic segmentation probability maps. However, since most of the existing methods calculate perceptual losses on an entire image in the same feature space, the results tend to be monotonous and unnatural. For this reason, Rad \etal~\cite{rad2019srobb} optimized SR models with a targeted objective function that penalizes images at different semantics using the corresponding terms. But, since the segmentation label needs to be fed to the SR network to calculate the targeted perceptual loss, the users cannot easily adjust the objective function. In summary, most early SR networks provide a designated HR output among many possible ones, not allowing us to explore more plausible outputs at the test phase. To alleviate this problem, Lugmayr \etal ~\cite{2020srflow} proposed the SRFlow using a normalizing flow method capable of learning the conditional distribution of the output given the low-resolution input. As a result, it can learn to predict diverse photo-realistic high-resolution images.
%------------------
Though great strides have been made, the natural and flexible reconstruction of local regions is still challenging. As stated previously, there can be diverse HR solutions for a given LR, meaning that one LR input can be restored to different HR results depending on the context and situation. Particularly because of various shapes and textures in the real world, the one-to-many problem becomes even more serious if the SR network's capacity is not large enough.
%------------------
To solve this problem, first, the SR model should be able to generate more diverse styles of HR reconstruction while keeping consistency with the given LR image. Second, the recovery style needs to be locally controlled. Third, training and storing too many redundant SR models with different parameters should be avoided. Achieving these requirements would enable us to explore various HR solutions for each region effectively. In this respect, some recent methods made it possible to continuously generate and adjust intermediate results between two objective functions, \ie perception and distortion functions ~\cite{wang2019cfsnet, wang2019deep, shoshan2019dynamic}. However, there can be some improvements in these approaches, as they defined just two objective functions and controlled the entire image, not the local regions needing adjustment.

%---------------------
%%%%%%%% figure 3 %%%%%%%% %%%%%%%% %%%%%%%% %%%%%%%% %%%%%%%% %%%%%%%% 

\begin{figure*}[!t]
  \centering
\begin{minipage}[t]{1.0\linewidth}
  \centerline{\includegraphics[width=0.85\linewidth]{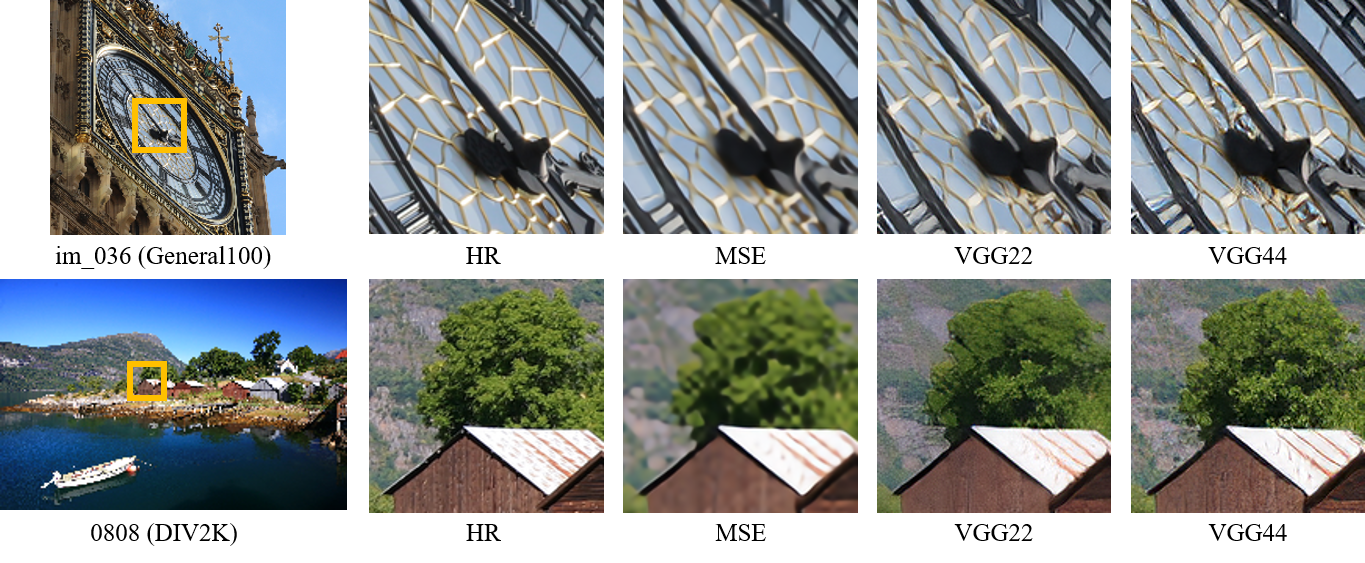}}
\end{minipage}
\caption{The effect of choosing different layers when estimating perceptual losses on different regions, e.g., on edge and texture regions, where the losses correspond to MSE, ReLU 2-2 (VGG22), and ReLU 4-4 (VGG44) of the VGG-19 network.}
\label{fig:f03}
\end{figure*}
%-------------------------

In this paper, we attempt locally adjustable HR generation by exploring the SR model optimization, focusing on the development of conditional objectives that can generate various reconstruction styles. The proposed objective consists of the weighted sum of several perceptual losses from different feature levels. The weights vary according to the condition, which is the recovery style information in our work. Experiments show that training an SR model with our multi-level perceptual losses generates various recovery styles effectively, which also enables us to finely control the styles of local regions.

%=======================
\section{Related Work}
\subsection{Loss Functions for SISR}
The choice of the objective function affects the recovery style and reconstruction performance. For instance, adversarial loss ~\cite{goodfellow2014generative} encourages an SR network to generate perception-oriented solutions~\cite{denton2015deep, xu2017learning, ren2020real, agustsson2017ntire}. Perceptual losses ~\cite{BrunaSL15, johnson2016perceptual} are proposed to optimize SR models by minimizing the error in the feature space instead of pixel space. Dovovitskiy \etal~\cite{dosovitskiy2016generating} and Ledig \etal~\cite{2017photo} proposed to use adversarial loss in combination with the perceptual loss to encourage the network to favor solutions that look more like natural images. With these loss functions, the overall visual quality of reconstruction is significantly improved ~\cite{Hui-PPON-2021, zhang2020ntire, tariq2020deep}. Recently, some studies ~\cite{lugmayr2019aim,jo2020investigating,yan2019deep} proposed to use GAN with losses based on perceptual quality assessment metric. Another perceptual loss is proposed in ~\cite{rad2019srobb}, using different levels of features according to semantic segmentation labels such as objects, boundaries, and backgrounds. In these approaches, once an SR model is trained, a fixed HR is produced for the LR input.
%-------------------------
\subsection{Network Conditioning}
The feature normalization techniques generally change networks' behavior based on the input properties. The representative normalization methods may be batch normalization (BN) ~\cite{ICML-2015-IoffeS} and instance normalization (IN) ~\cite{UlyanovVL16}. The IN normalizes a single image while the BN does a whole batch of images. Conditional Instance Normalization (CIN) has also been introduced in ~\cite{DumoulinSK16}, which uses the learned representations to model multiple styles simultaneously. Huang \etal~\cite{huang2017arbitrary} proposed adaptive instance normalization (AdaIN) to adjust features to arbitrary new styles. Perez \etal~\cite{perez2018film} proposed Feature-wise Linear Modulation, called FiLM, as a general-purpose conditioning method for neural networks. FiLM layers influence neural network computation via a simple, feature-wise affine transformation based on conditioning information. Inspired by these works, Wang \etal~\cite{2018recovering} proposed a spatial feature transformation (SFT) layer to modulate the features of some intermediate layers in a single network conditioned on semantic segmentation probability maps.
Our approach is partially inspired by the above feature normalization methods, which can alter the behavior of deep CNNs to influence the output. In terms of network architecture, we use the Residual-in-Residual Dense Block (RRDB) ~\cite{2018esrgan} equipped with SFT layers.
%-------------------------
\subsection{Continuous Imagery Effect Transition}
Since the restored image's perceived quality is relatively subjective, and the perception-oriented methods sometimes generate artifacts, users may wish to control the reconstruction result according to the preferences or image characteristics. In recent years, there have been some tunable models that produce intermediate images between the goals of two different objective functions. Specifically, these methods start by training several separate models and then propose different ways of interpolating between them, specifically by directly interpolating the output pixels or network weights ~\cite{2018esrgan, wang2019deep}, or by using specialized adaptor blocks in the networks~\cite{shoshan2019dynamic}. They considered trade-off relationships between two objectives, such as perception-distortion balance in SR, noise reduction vs. detail preservation in denoising and style transfer~\cite{he2019modulating,shoshan2019dynamic, wang2019cfsnet, wang2019deep}. However, these methods have some limitations: the number of objective functions is two, and they cannot adjust local regions, \ie the algorithm is equally applied to the entire region of an image. It is also inefficient that they have to train and store multiple separate models. On the other hand, Bahat \etal~\cite{bahat2020explorable} proposed an explorable SR framework that enables local restoration control. However, users have to manually edit the texture in a few steps through a user interface. For easier and more effective quality control, we propose a controllable SR model that can produce various recovery styles for each region with a simple adjustment method. Besides, we can generate intermediate results between two or more different styles at fine control levels.
%----------------------------
\subsection{Multi-task Learning}
Learning one task at a time is a typical methodology in machine learning because it is hard to simultaneously optimize multiple objectives due to model capacity limitation or conflicting losses. For this reason, such multi-objective problems are commonly scalarized by a linear combination of the losses, with weights defining the trade-off between the loss term ~\cite{caruana1997multitask}. On the other hand, Multi-task Learning (MTL) is an inductive transfer mechanism whose goal is to improve generalization performance by leveraging useful domain-specific information contained in multiple related tasks ~\cite{baxter2000model}. Specifically, since the MTL networks use shared layers trained in parallel on all the tasks, what is learned for each task can help others to learn better when tasks are closely related  ~\cite{caruana1997multitask, zhang2021survey}. Recently, Dosovitskiy \etal ~\cite{dosovitskiy2019you} proposed loss-conditional training of deep networks for MTL that can improve model efficiency by exploiting the redundancy of multiple related models. They demonstrate style-transfer trained in this way and utilize feature-wise linear modulation ~\cite{perez2018film} that affects the whole image style.

%%%%%%%% figure - f04
\begin{figure*}[!t]
  \centering
\begin{minipage}[t]{1.0\linewidth} 
  \centerline{\includegraphics[width=0.85\linewidth]{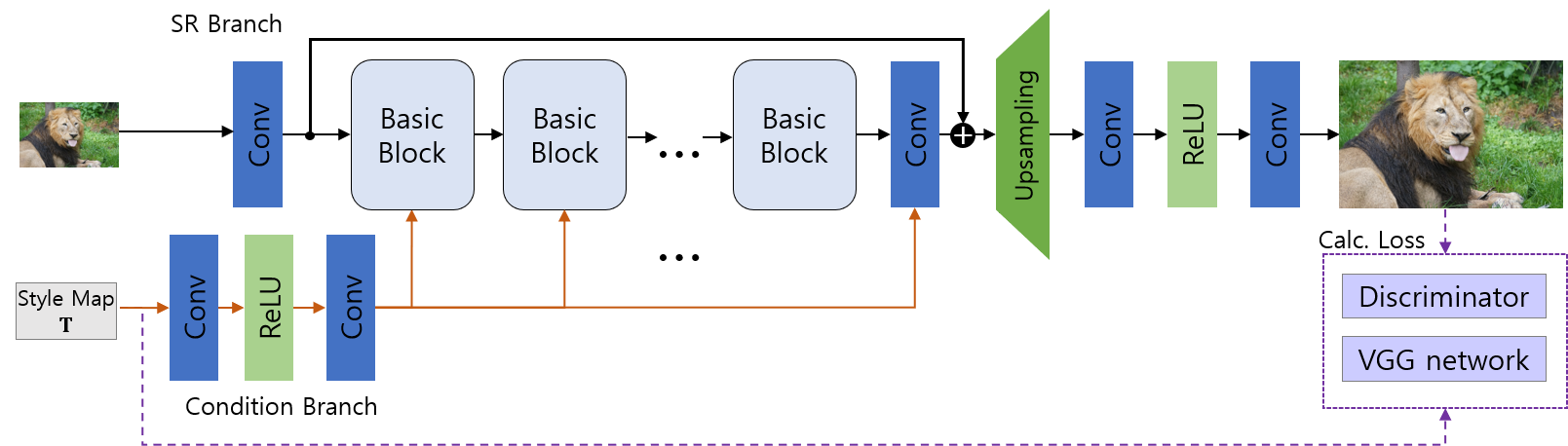}}
\end{minipage}
\caption{The architecture of our proposed flexible SR network. We use the RRDB equipped with SFT as a basic block (Figure~\ref{fig:f05}(c)). The condition branch takes a style map for reconstruction style as input. This map is used to control the recovery styles of edges and textures for each region through SFT layers.}
\label{fig:f04}
\end{figure*}

%============================
\section{Proposed Method}
\subsection{Targeted Perceptual Loss}
In general, the choice of feature space significantly influences perceptual reconstruction performance and the styles. For example, Figure~\ref{fig:f03} shows the effect of choosing different feature spaces in computing the perceptual loss. In this paper, four different layers, ReLU 2-2, ReLU 3-4, ReLU 4-4, and ReLU 5-4 of the VGG-19 network~\cite{SimonyanZ14a} are considered, denoted as VGG22, VGG34, VGG44, and VGG54, respectively. As shown in Figure~\ref{fig:f03}, while the low-level feature space VGG22 seems more suitable for reconstructing simple edges with less distortion and over-sharpening, the mid- and high-level feature spaces of VGG44 are more appropriate for recovering complex textures. Therefore, it is difficult to determine a single feature space that works best for the entire image.
%----------------------------
In our work, we use more than two feature spaces at the same time to train a flexible SR (FxSR) model capable of generating various reconstruction styles. We define two kinds of FxSR models, namely FxSR-PD (perception-distortion) and FxSR-DS (diversity). The FxSR-PD is the main model in our work, which controls the output style between the distortion-oriented and perception-oriented by combining the reconstruction loss (for distortion) and VGG22 feature loss (for perception), along with the adversarial loss. The FxSR-DS uses the same architecture as the FxSR-PD but is trained with different losses, including all the VGG features stated above. Hence, the aim of FxSR-DS is to produce diverse styles of outputs related to different VGG features rather than to control between distortion and perception. Unlike previous works where there is no control data, we adjust the network by applying different objective functions for each local region through a style control map\footnote{In the rest of the paper, we will refer the style control map as just {\em style map} or a map T.}. As a result, we can explore various HR solutions that are generated using multiple objective functions and thus reconstruct an image with the desired style or an image closer to the original HR.
%----------------------------
%%%%%%%% figure 5  with SFT for basic blocks
\begin{figure}[!t]
\centering
\scriptsize
\begin{minipage}[t]{1.0\linewidth}
    \centering
    \subfigure[RB with SFT ~\cite{2018recovering}] {\includegraphics[width=0.19\linewidth
    ]{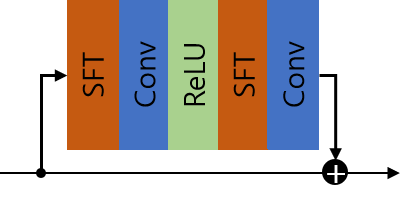}}\hfill
    \subfigure[Residual-in-Residual Dense Block (RRDB) ~\cite{2018esrgan}] {\includegraphics[width=0.69\linewidth
    ]{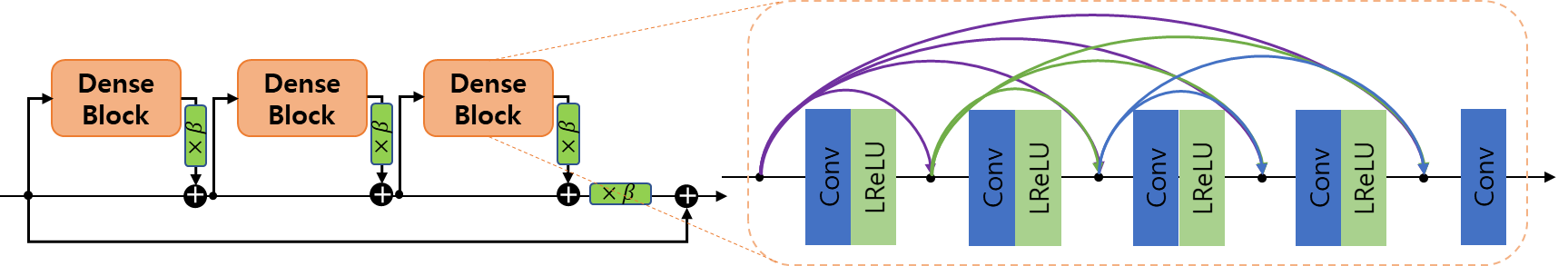}}\vfill
    \subfigure[The proposed Basic Block (RRDB equipped with SFT layer)]
    {\includegraphics[width=0.70\linewidth
    ]{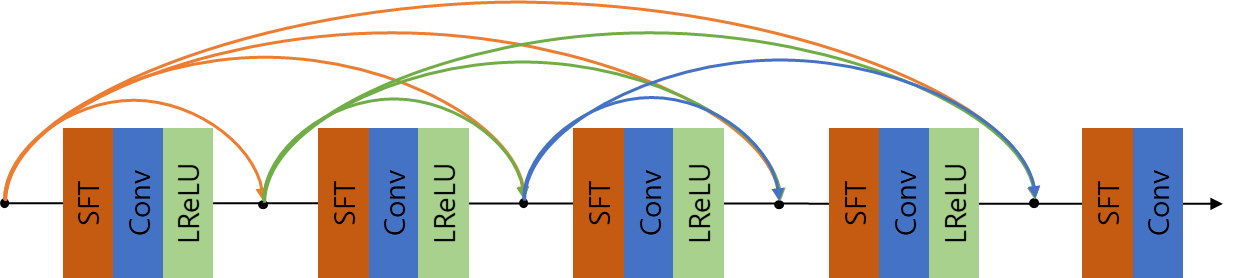}}\vfill
\end{minipage}
\caption{RRDB with SFT for basic blocks}
\label{fig:f05}
\end{figure}

\subsection{Proposed SR with Flexible Style}
Given a single LR image ${I}^{LR}$, SISR is to estimate an HR image $\hat{I}^{HR}$, which is as similar as possible to its corresponding HR counterpart ${I}^{HR}$. Most of the current CNN-based methods use feed-forward networks to directly learn a mapping function ${G}_{\theta}$ parameterized by ${\theta}$ as
\begin{equation}
\hat{I}^{HR}={G}_{\theta}\left({I}^{LR}\right).
\label{eqn:predHR}
\end{equation}
To optimize ${G}_{\theta}$ on the training samples, we design a specific objective function $\mathcal{O}$ as
\begin{equation}
{\theta}^\ast=\arg\min_{\theta}\mathbb{E}_{\mathcal{Z}\sim\mathcal{P}_{\mathcal{Z}}}\left[\mathcal{O}\left(\hat{I}^{HR},{I}^{HR}\right)\right]
\label{eqn:optTheta}
\end{equation}
where $\mathcal{Z}=\left({I}^{LR},{I}^{HR}\right)$ is sampled from given a training distribution of pairs $P_{\mathcal{Z}}$. Many recent studies~\cite{2017photo, sajjadi2017enhancenet} use perceptual loss and adversarial loss for designing $\mathcal{O}$ to recover realistic textures. Although these losses greatly improve the perceptual quality, the generated textures tend to be monotonous and unnatural ~\cite{rad2019srobb, 2018recovering}. To further improve the restoration performance, Wang \etal~\cite{2018recovering} used semantic segmentation probability maps as the categorical prior $\Psi$ and reformulated (\ref{eqn:predHR}) as
\begin{equation}
\hat{I}_{\Psi}^{HR}=G_{\theta}\left({I}^{LR}\right|\Psi ).
\label{eqn:predHRwPsi}
\end{equation}

%----------------------------

However, the perceptual loss was applied to the entire region of images, like in previous works. Specifically, the same level of features was used both on simple edges and complex textures, which has a limitation in restoring images composed of various types of objects.
In addition, once model training is completed, there is no way to adjust the SR results without retraining. Hence, instead, we propose a novel method to apply different objectives to each region for reconstructing desired images or images closer to the original. Specifically, the proposed flexible SR model is optimized with a conditional objective, which is a weighted sum of several perceptual losses corresponding to different feature levels, where each weight changes depending on the style map. 
%----------------------------
Formally, our objective is described as:
\begin{equation}
\hat{I}_\mathbf{T}^{HR}={G}_{\theta}\left({I}^{LR}\right|\mathbf{T}),
\label{eqn:predHRwT}
\end{equation}

\begin{equation}
{\theta}^\ast=\arg\min_{\theta}\mathbb{E}_{{t}\sim\mathcal{P}_{t}}\mathbb{E}_{\mathcal{Z}\sim\mathcal{P}_{\mathcal{Z}}}\left[\mathcal{O}\left(\hat{I}^{HR}_\mathbf{T},{I}^{HR}|\mathbf{T}\right)\right]
\label{eqn:optThetawT}
\end{equation}
where $\mathbf{T}$ is a map delivering spatially varying style control. That is, the map $\mathbf{T}$ is an LR-sized matrix, which is fed to the condition network to change the SR styles. Since the purpose of training is to let the network learn various styles corresponding to given control parameters, we feed various $\mathbf{T}$ randomly to the network during the training. Specifically, we feed a flat map $\mathbf{T} =t \times \mathbf{1}$ during the training, where $\mathbf{1}$ is the matrix with all the elements 1, and $t$ is a variable related to the feature combinations, which will be detailed in the following subsection. For training with various feature combinations, we change $t$ randomly at each epoch. At the inference, if we feed a flat map as defined above, the network will deliver an SR style globally corresponding to the $t$. If we wish to control the styles locally, we feed a spatially varying map, which will be demonstrated in the experiment.
%----------------------------
\subsection{Proposed Network Architecture}
An overview of the architecture is shown in Figure~\ref{fig:f04}. The generator network ${G}_{\theta}$ consists of two streams, an SR branch and a condition branch. The SR branch is built with basic blocks consisting of RRDB equipped with the SFT layers~\cite{2018recovering}, which take the shared conditions as input and modulate feature maps by applying the affine transformation. This structure is shown in Figure~\ref{fig:f05}(c), where the residual block with SFT~\cite{2018recovering} and RRDB~\cite{2018esrgan} are also shown in Figures~\ref{fig:f05}(a) and (b) for comparison. The SFT layer learns a mapping function that outputs a modulation parameter based on a style condition $\mathbf{T}$. This modulation layer allows the SR branch to optimize the changing objective during the training and also to generate SR results with spatially different styles according to the style map. The condition branch is used to produce shared intermediate style conditions that can be broadcasted to all the SFT layers for efficiency. As in the study of~\cite{2018recovering}, all the convolution layers in the condition branch are restricted to use ${1}\times{1}$ kernels to avoid the interference of different regions. For discriminator network, we use VGG network~\cite{SimonyanZ14a} that contains ten convolution layers gradually decreasing the spatial dimensions.
%----------------------------
%%%%%%%% figure fig_FxSR_w 
\begin{figure}[!t]
\centering
% \scriptsize
\begin{minipage}[t]{1.0\linewidth}
    \centering
    {\includegraphics[width=0.7 \linewidth
    ]{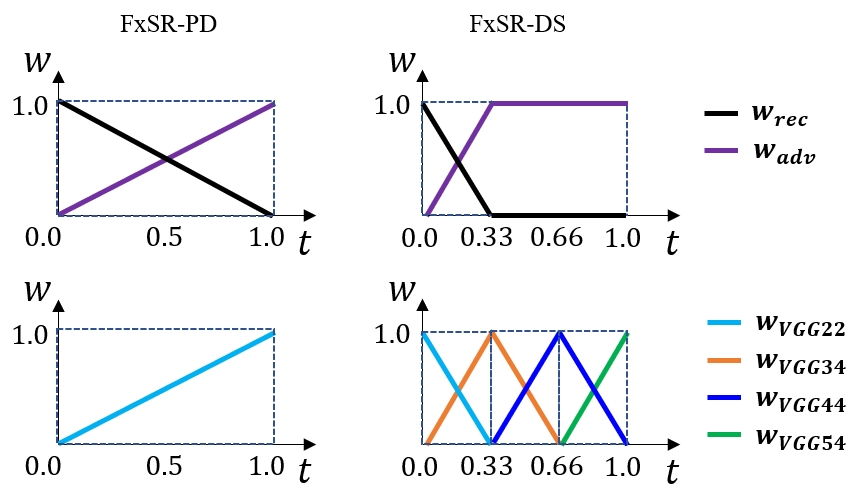}}\vfill
\end{minipage}
\caption{The left column shows the weight functions for FxSR-PD. $t=0$ corresponds to distortion-oriented SR (only MSE loss) and $t=1$ perception-oriented (with adversarial and perceptual loss from VGG22). The right column shows the weight functions for FxSR-DS, where more perceptual losses are used to expand the HR styles.}
\label{fig:FxSR_w}
\end{figure}
%----------------------------
\subsection{Proposed Loss Function}
We combine multiple losses to train our SR model. The conditional objective consists of three terms, namely pixel-wise reconstruction loss, adversarial loss, and proposed conditional perceptual loss:

\begin{equation}
\mathcal{O}_\mathbf{T}={\lambda}_{rec}(\mathbf{T})\cdot\mathcal{L}_{rec}+{\lambda}_{adv}(\mathbf{T})\cdot\mathcal{L}_{adv}+{\lambda}_{per}\cdot\mathcal{L}_{per}(\mathbf{T})
\end{equation}
where
\begin{equation}
{\lambda}_{rec}(\mathbf{T})={\lambda}_{rec\_o}+(\eta\cdot{w}_{rec}(\mathbf{T})),
\end{equation}
\begin{equation}
{\lambda}_{adv}(\mathbf{T})={\lambda}_{adv\_o}\cdot{w}_{adv}(\mathbf{T}).
\end{equation}
The notations will be explained one by one below.
First, the reconstruction loss is calculated as:
\begin{equation}
\mathcal{L}_{rec}=\mathbb{E}\left[\lVert\hat{I}^{HR}-{I}^{HR}\rVert_{1}\right].
\end{equation}
We use the adversarial loss using Relativistic average Discriminator RaD~\cite{Jolicoeur-Martineau19} that performs better for learning sharper edges and more detailed textures compared to standard GAN~\cite{goodfellow2014generative}. While the standard version estimates the probability that one input image $I$ is real and natural, the RaD predicts the probability that a real image ${I}^{HR}$ is relatively more realistic than a fake one $\hat{I}^{HR}$. In addition, for adversarial training, RaD benefits from the gradients from both $\hat{I}^{HR}$ and ${I}^{HR}$, while only $\hat{I}^{HR}$ takes effect in the standard version.
%----------------------------
Specifically, the adversarial and the discriminator losses are:
\begin{equation}
\footnotesize
\mathcal{L}_{adv}=-\mathbb{E}_{\hat{I}^{HR}}\left [ \mathrm{log}\left ( \widetilde{D}\left (\hat{I}^{HR}\right ) \right ) \right ]-\mathbb{E}_{{I}^{HR}}\left [ \mathrm{log}\left (1- \widetilde{D}\left ( {I}^{HR} \right ) \right ) \right ]
\end{equation}
\begin{equation}
\footnotesize
\mathcal{L}_{dis}=-\mathbb{E}_{I^{HR}}\left [ \mathrm{log}\left ( \widetilde{D}\left ( I^{HR} \right ) \right ) \right ]-\mathbb{E}_{\hat{I}^{HR}}\left [ \mathrm{log}\left (1- \widetilde{D}\left (\hat{I}^{HR}\right ) \right ) \right ]
\end{equation}
where 
\begin{equation}
\footnotesize
\widetilde{D}\left ( I^{HR} \right ) = \mathrm{sigmoid} \left ( C\left (  I^{HR}\right ) -\mathbb{E}_{\hat{I}^{HR}}\left [ C\left (\hat{I}^{HR} \right )\right ] \right )
\end{equation}
\begin{equation}
\footnotesize
\widetilde{D}\left ( \hat{I}^{HR} \right ) = \mathrm{sigmoid} \left ( C\left (  \hat{I}^{HR}\right ) -\mathbb{E}_{{I}^{HR}}\left [ C\left ( {I}^{HR} \right )\right ] \right )
\end{equation}
where $C\left ( \cdot  \right )$ represents the output logit of discriminator.
%----------------------------
The conditional perceptual loss is a weighted sum of multiple perceptual losses in different levels of feature spaces:
\begin{equation}
\mathcal{L}_{per}(\mathbf{T})=\sum_{l}{w}_{l}(\mathbf{T})\cdot\mathcal{L}_{l},
\end{equation}
where $\mathcal{L}_{l}$ denotes the distance in each feature space, $l\in \{VGG12$,$VGG22,$ $\cdots,VGG54\}$, and the weights ${w}_{l}$ changes according to $\mathbf{T}$.
Precisely, the distance $\mathcal{L}_{l}$ is defined as
\begin{equation}
\mathcal{L}_{l}=\mathbb{E}\left[\lVert{\phi_{l}(\hat{I}^{HR})-\phi_{l}({I}^{HR}))}\rVert_{2}\right]
\end{equation}
where ${\phi}_{l}$ denotes feature maps in the feature space $l$.
The weights ${w}_{rec}$, ${w}_{adv}$, and ${w}_{l}$ are functions of $t$ as described in Figure~\ref{fig:FxSR_w}, where $t$ is a random variable having uniform distribution in $[0,1]$ during the training.
%----------------------------
\subsection{Implementation details}
This subsection explains how we design the combination of feature losses depending on the change of $t$.
The left column of Figure~\ref{fig:FxSR_w} shows the weight function for FxSR-PD (using only VGG22 for perceptual loss), and the right for FxSR-DS (using more feature spaces for diversity).
When $t$=0, the figure shows that FxSR-PD corresponds to distortion-oriented SR (perceptual and adversarial losses are zero). When the value of $t$ approaches 1, then it becomes perception-oriented (weight for the reconstruction loss becomes zero, while adversarial and perceptual losses grow to 1). In the case of the right column, various feature distances are involved in the perceptual loss, and hence FxSR-DS can deliver diverse styles. Specifically, note that $t=1$ corresponds to a perception-oriented SR with VGG54 as the feature space. Also, even when $t$ approaches 0, the FxSR-DS still produces perception-oriented SR results of different styles corresponding to VGG22, unlike the FxSR-PD that is distortion-oriented at $t=0$.
%----------------------------
Regarding the style control, as stated previously, we use a uniform map $\mathbf{T} =t \times \mathbf{1}$ at the training phase. That is, a flat map is fed to the condition branch, with its intensity $t$ randomly changing during the training. Since the SR network is a fully convolutional neural network, it inherits the local connectivity property that the local image and the map region determine the output pixel. Hence, SR models trained with uniform maps can handle spatially varying cases. 
%----------------------------

%%%%%%%% FxSR-PD T %%%%%%%% %%%%%%%% 
\begin{figure*}[!t]

\setlength{\arrayrulewidth}{1.0pt}
\newcolumntype{Z}
{>{\centering\arraybackslash}X}
\begin{center}
\small
%\footnotesize

\renewcommand{\tabcolsep}{1pt}
\begin{tabularx}{\linewidth}{Z Z Z Z Z Z}
\hline  
      Whole image & HR & \multicolumn{4}{c}{$4\times$ FxSR-PD} \\
\hline
\\
&&\multicolumn{4}{c}{\includegraphics[width=0.55\linewidth]{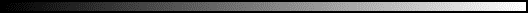}}\\
        &  & $t=0.0$ & $t=0.3$ & $t=0.6$ & $t=1.0$ \\
% \hline
\end{tabularx}
\end{center}

\centering
% \scriptsize
\begin{minipage}[t]{1.0\linewidth}
    \centering
    
    \includegraphics[width=0.16\linewidth]{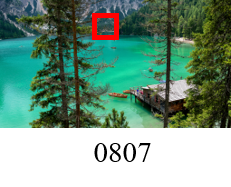}\hfill
    \includegraphics[width=0.16\linewidth]{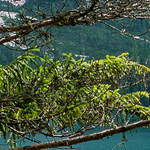}\hfill
    \includegraphics[width=0.16\linewidth]{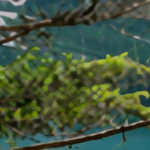}\hfill
    \includegraphics[width=0.16\linewidth]{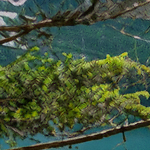}\hfill
    \includegraphics[width=0.16\linewidth]{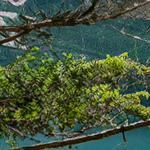}\hfill
    \includegraphics[width=0.16\linewidth]{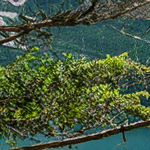}\vfill
    \vspace{0.3cm}

    \includegraphics[width=0.16\linewidth]{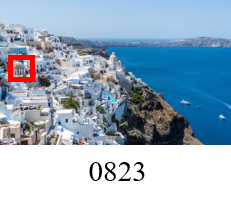}\hfill
    \includegraphics[width=0.16\linewidth]{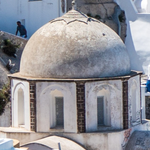}\hfill
    \includegraphics[width=0.16\linewidth]{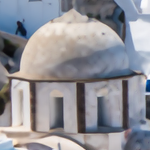}\hfill
    \includegraphics[width=0.16\linewidth]{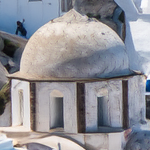}\hfill
    \includegraphics[width=0.16\linewidth]{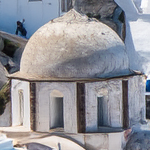}\hfill
    \includegraphics[width=0.16\linewidth]{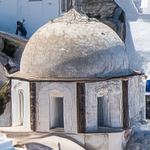}\vfill
    \vspace{0.3cm}

    \includegraphics[width=0.16\linewidth]{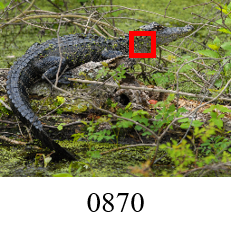}\hfill
    \includegraphics[width=0.16\linewidth]{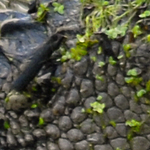}\hfill
    \includegraphics[width=0.16\linewidth]{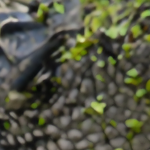}\hfill
    \includegraphics[width=0.16\linewidth]{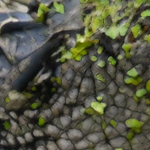}\hfill
    \includegraphics[width=0.16\linewidth]{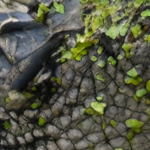}\hfill
    \includegraphics[width=0.16\linewidth]{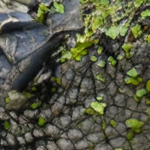}\vfill
    \vspace{0.3cm}
    
    \includegraphics[width=0.16\linewidth]{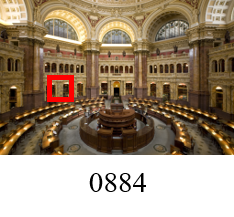}\hfill
    \includegraphics[width=0.16\linewidth]{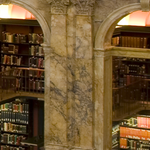}\hfill
    \includegraphics[width=0.16\linewidth]{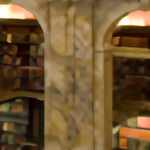}\hfill
    \includegraphics[width=0.16\linewidth]{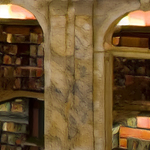}\hfill
    \includegraphics[width=0.16\linewidth]{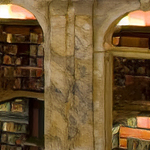}\hfill
    \includegraphics[width=0.16\linewidth]{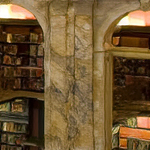}\vfill
    \vspace{0.3cm}
    
    \includegraphics[width=0.16\linewidth]{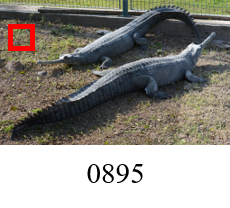}\hfill
    \includegraphics[width=0.16\linewidth]{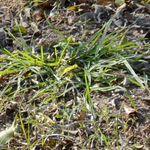}\hfill
    \includegraphics[width=0.16\linewidth]{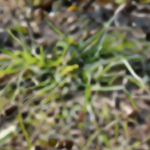}\hfill
    \includegraphics[width=0.16\linewidth]{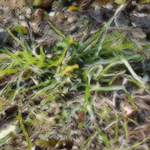}\hfill
    \includegraphics[width=0.16\linewidth]{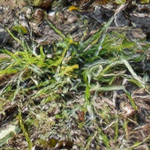}\hfill
    \includegraphics[width=0.16\linewidth]{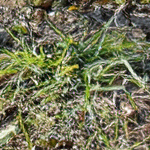}\vfill
    
\end{minipage}
\caption{Changes in the result of FxSR-PD $4\times$ SR according to $t$ on DIV2K validation set~\cite{agustsson2017ntire} .}
\label{fig:fig_FxSR-PD_t}
\end{figure*}

\section{Experiments}
In the experiment, we compare our FxSR-PD and FxSR-DS with several state-of-the-art SR methods on benchmark datasets. We start the section with a description of the datasets and evaluation methods. Next, we present the comparison results. We also provide examples of local style control and validate the effectiveness of our approach for compressed images. Finally, we report complexity analysis for the proposed methods.
% \hl{In this section, we compare the proposed FxSR model with several
% state-of-the-art SR methods on benchmark datasets. We start the section with a description of the datasets and performance metrics used for the evaluation. Next, we present comparative results of FxSR-PD and FxSR-DS with the state-of-the-art methods and report complexity analysis. We also provide examples of per-pixel style control and validate the effectiveness of our method for compressed inputs.}

\subsection{Materials and Methods} 

\subsubsection{Datasets}
For the experiments, we train the FxSR with DIV2K ~\cite{agustsson2017ntire} dataset, which contains 800 training images, 100 validation images, and 100 test images. We use BSDS100, General100, and DIV2K 100 validation images as our test datasets. We also use JPEG-compressed images for training and testing FxSR models to show that our proposed method is still effective on the real-world compressed LR images. The scaling factors of ${4}\times$ and ${8}\times$ are tried for experiments.

%%%%%%%% figure PSNR 4x %%%%%%%% %%%%%%%% %%%%%%%% 
\begin{figure*}[!t]
\centering
\scriptsize
\begin{minipage}[t]{1.0\linewidth}
    \centering
    \subfigure[] {\includegraphics[width=0.32\linewidth
    ]{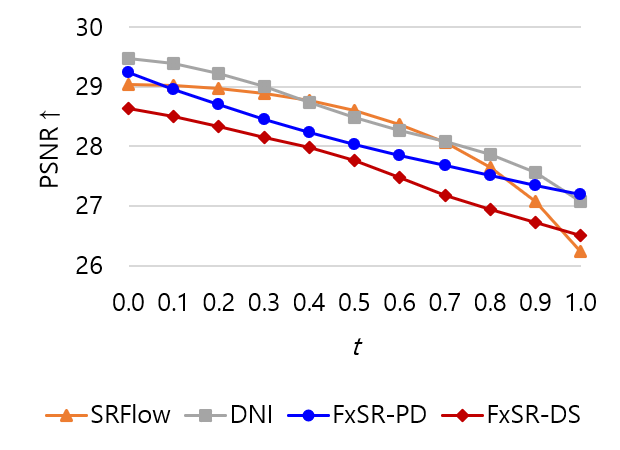}}\hfill
    \subfigure[] {\includegraphics[width=0.32\linewidth
    ]{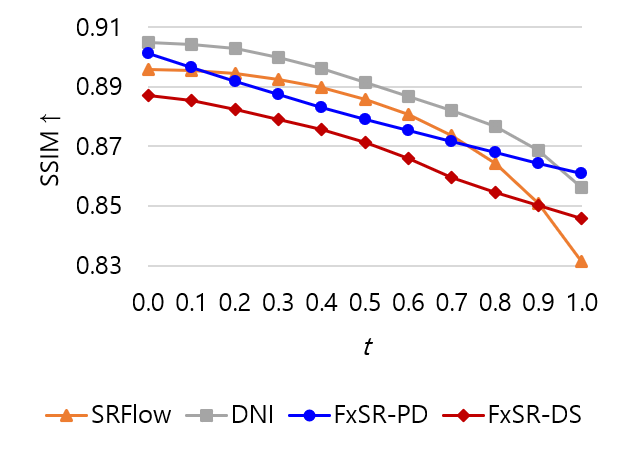}}\hfill
    \subfigure[] {\includegraphics[width=0.32\linewidth
    ]{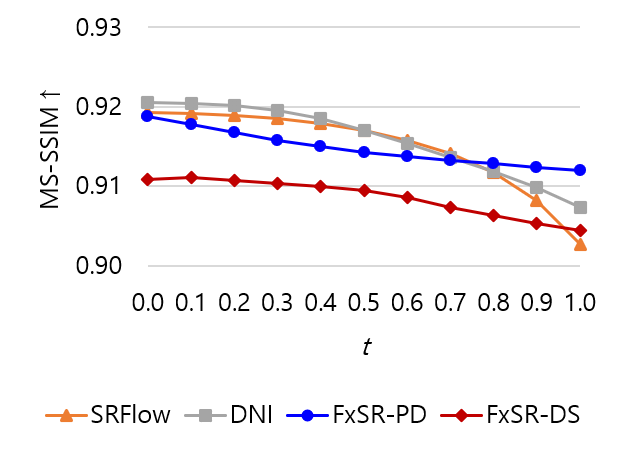}}\hfill
\end{minipage}
\caption{${4}\times$ SR performance comparison of state-of-the-art and proposed methods evaluated by the metric (a) PSNR, (b) SSIM ~\cite{wang2004image}, and (c) MS-SSIM ~\cite{wang2003multiscale} for DIV2K according to condition parameters.}
\label{fig:fig_PSNR_4x}
\end{figure*}

%%%%%%%% figure LPIPS 4x %%%%%%%% %%%%%%%% %%%%%%%% 
\begin{figure*}[!t]
\centering
\scriptsize
\begin{minipage}[t]{1.0\linewidth}
    \centering
    \subfigure[] {\includegraphics[width=0.32\linewidth
    ]{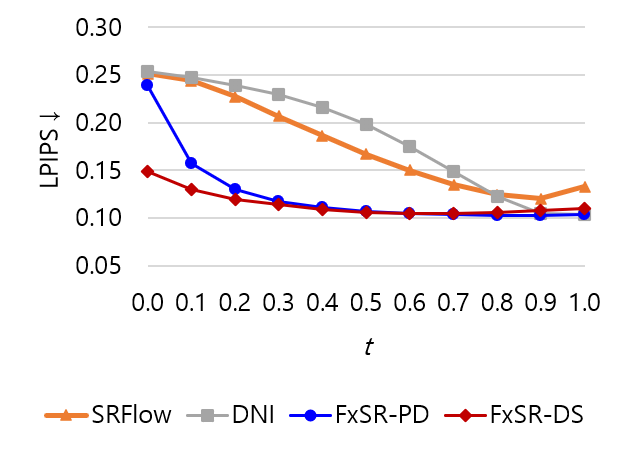}}\hfill
    \subfigure[] {\includegraphics[width=0.32\linewidth
    ]{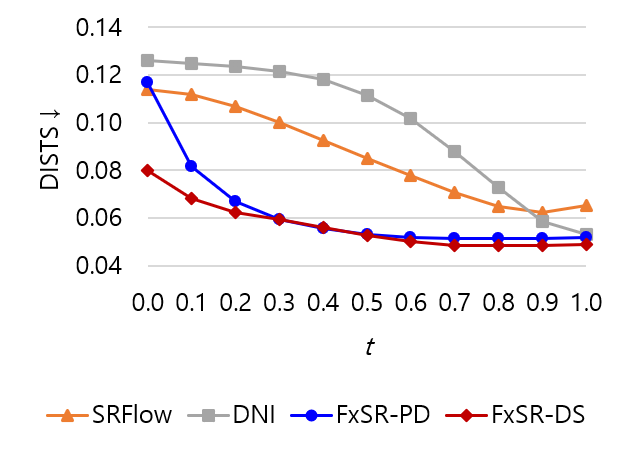}}\hfill
    \subfigure[] {\includegraphics[width=0.32\linewidth
    ]{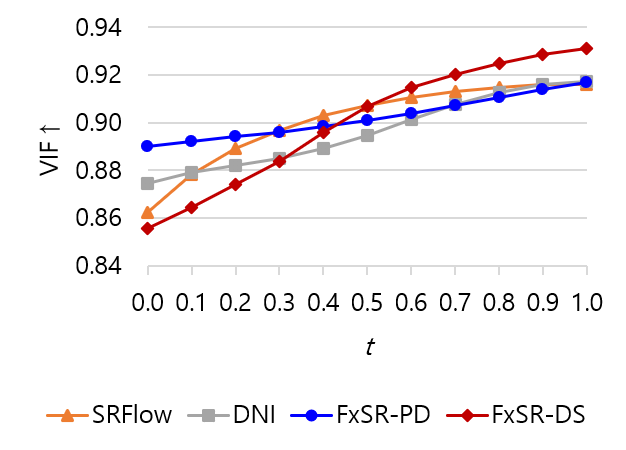}}\hfill
\end{minipage}
\caption{${4}\times$ SR performance comparison of state-of-the-art and proposed methods evaluated by the metric (a) LPIPS ~\cite{zhang2018unreasonable}, (b) DISTS ~\cite{ding2020image}, and (c) VIF ~\cite{sheikh2006image} for DIV2K according to condition parameters.}
\label{fig:fig_LPIPS_4x}
\end{figure*}

%%%%%%%% figure NIQE 4x %%%%%%%% %%%%%%%% %%%%%%%% 
\begin{figure*}[!t]
\centering
\scriptsize
\begin{minipage}[t]{1.0\linewidth}
    \centering
    \hfill
    \subfigure[] {\includegraphics[width=0.32\linewidth
    ]{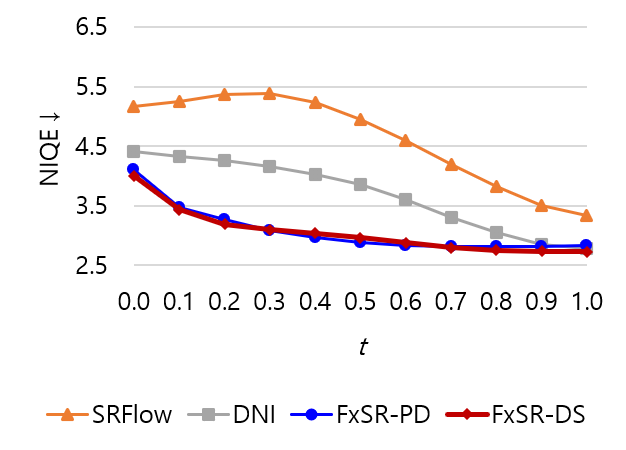}}\hfill
    \subfigure[] {\includegraphics[width=0.32\linewidth
    ]{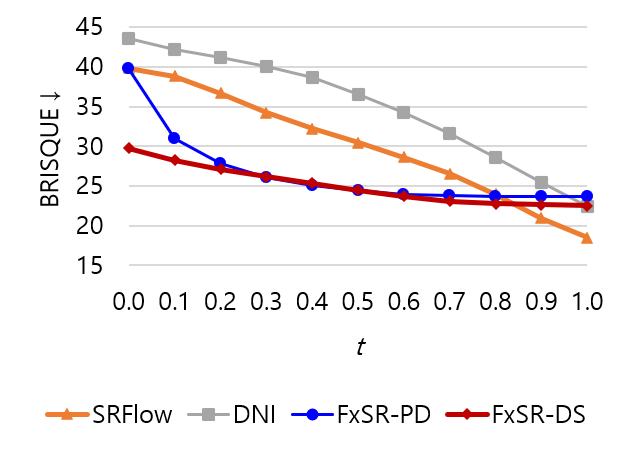}}\hfill\hfill
    % \subfloat[] {\includegraphics[width=0.32\linewidth
    % ]{figure/fig_comp_VIF_4x.png}}\hfill
\end{minipage}
\caption{${4}\times$ SR performance comparison of state-of-the-art and proposed methods evaluated by the (a) NIQE ~\cite{mittal2012making} and (b) BRISQUE ~\cite{mittal2012no} for DIV2K according to condition parameters.}
\label{fig:fig_NIQE_4x}
\end{figure*}
%%%%%%%% figure 1 %%%%%%%% %%%%%%%% %%%%%%%% 
\begin{figure*}[!t]
  \centering
\begin{minipage}[t]{0.90\linewidth}
    \centerline{\includegraphics[width=0.80\linewidth]{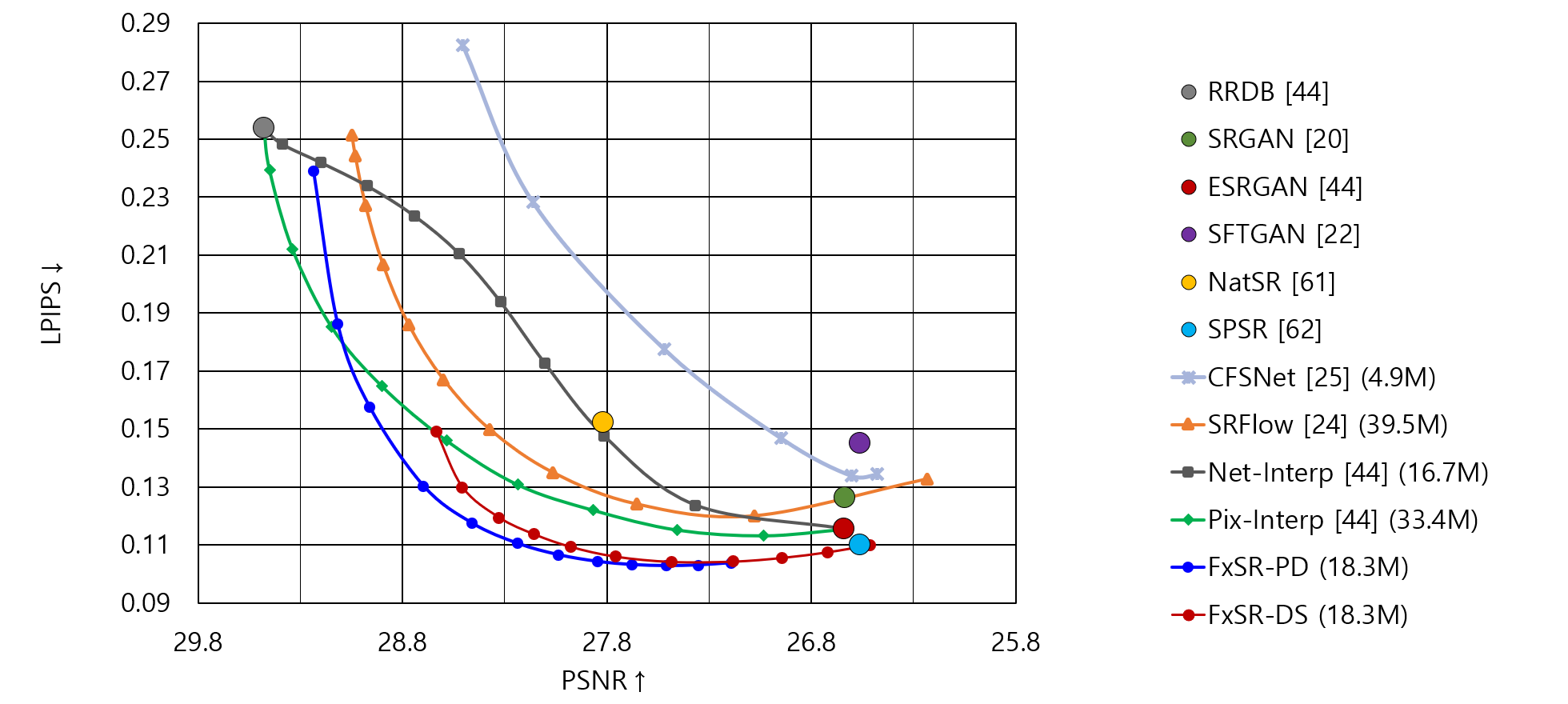}}\vfill
\end{minipage}
\caption{Performance comparison of the state-of-the-arts and proposed method
(FxSR-PD: Perception-Distortion Flexible SR) for DIV2K {4}$\times$ SR. 
}
\label{fig:f01}
\end{figure*}
%%%%%%%% figure P-D 4x %%%%%%%% %%%%%%%% %%%%%%%% 
\begin{figure*}[!t]
\centering
\scriptsize
\begin{minipage}[t]{0.9\linewidth}
    \centering
    \subfigure[] {\includegraphics[width=0.32\linewidth
    ]{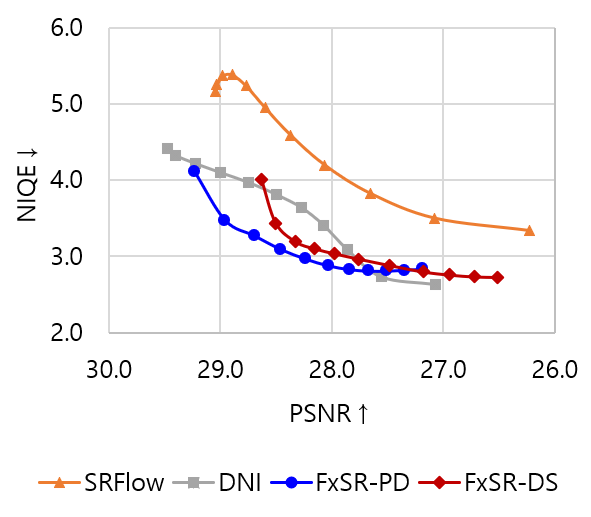}}\hfill
    \subfigure[] {\includegraphics[width=0.32\linewidth
    ]{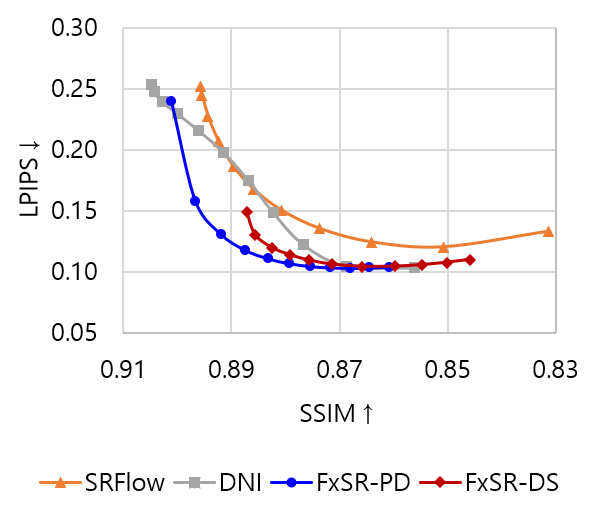}}\hfill
    \subfigure[] {\includegraphics[width=0.32\linewidth
    ]{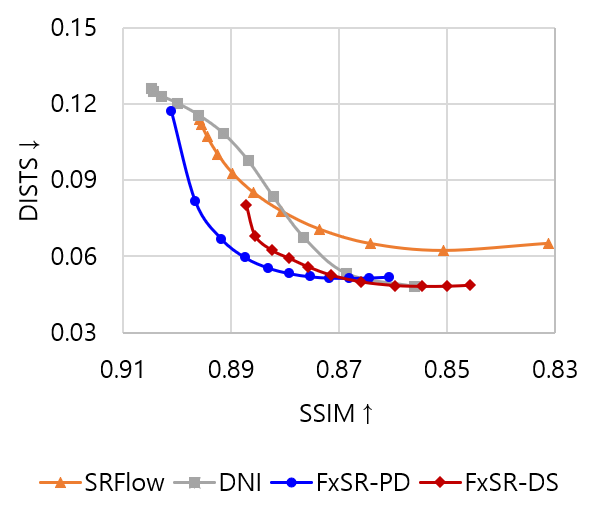}}\hfill
\end{minipage}
\caption{${4}\times$ SR P-D performance comparison of state-of-the-art and proposed methods evaluated by the metric (a) NIQE ~\cite{mittal2012making} Vs. PSNR, (b) SSIM ~\cite{wang2004image} Vs. LPIPS ~\cite{zhang2018unreasonable}, and (c) SSIM ~\cite{wang2004image} Vs. DISTS ~\cite{ding2020image} for DIV2K according to condition parameters.}
\label{fig:fig_P-D_4x}
\end{figure*}
%%%%%%%%% TABLE 1 - opt 2 %%%%%%%% %%%%%%%% 
\begin{table*}[!b]
\caption{Comparison with state-of-the-art SR methods on benchmarks. In {4}$\times$ and {8}$\times$, the 1st and the 2nd best performances are highlighted in \textcolor{red}{red} and \textcolor{blue}{blue}, respectively.}
% \vspace{-0.3cm}
\newcolumntype{Z}
{>{\centering\arraybackslash}X}

\begin{center}
\scriptsize
% \tiny
\renewcommand{\tabcolsep}{1pt}
%\footnotesize
%\small
\begin{tabularx}{\linewidth}{|Z|Z||Z||Z|Z|Z|Z|Z|Z|Z|Z|Z||Z||Z|Z|Z|Z|Z|Z|}
\hline
    &&\multicolumn{10}{c||}{{4}$\times$}&\multicolumn{6}{c|}{{8}$\times$}\\
    % &&    &&&&&{4}$\times$&&&&         &&&&{8}$\times$&&\\ 
\hline
\hline
    \multirow{3}*{Dataset} &  \multirow{3}*{Metric}  & { RRDB } ~\cite{2018esrgan}   & SRGAN ~\cite{2017photo}  & {ESR- GAN} ~\cite{2018esrgan} & {SFT- GAN} ~\cite{2018recovering} & {NatSR} ~\cite{2019natural}  & {SPSR} ~\cite{2020structure}   & SRFlow $\tau$=0.0 ~\cite{2020srflow} & SRFlow $\tau$=0.9 ~\cite{2020srflow} & FxSR-PD $t$=0.0 & FxSR-PD $t$=0.8 & { RRDB } ~\cite{2018esrgan} & {ESR- GAN} ~\cite{2018esrgan} & SRFlow $\tau$=0.0 ~\cite{2020srflow} & SRFlow $\tau$=0.9 ~\cite{2020srflow} & FxSR-PD $t$=0.0 & FxSR-PD $t$=0.8\\ 
\hline\hline

BSD& PSNR$\uparrow$ & \textcolor{red}{26.53} & 24.13 & 23.95 & 24.09 & 25.13 & 24.16 & 26.23 & 24.66 & \textcolor{blue}{26.38} & 24.77  &  \textcolor{blue}{23.56} & 20.23 & 23.37 & 21.66 & \textcolor{red}{23.60} & 21.93 \\

100& SSIM$\uparrow$ & \textcolor{red}{0.7438} & 0.6454 & 0.6463 & 0.6460 & 0.6780 & 0.6531 & 0.7293 & 0.6580 & \textcolor{blue}{0.7380} & 0.6817 & \textcolor{red}{0.5700} & 0.4350 & 0.5428 & 0.4632 & \textcolor{blue}{0.5728} & 0.5039 \\

& \tiny{LRPSNR}$\uparrow$ & \textcolor{blue}{51.52} & 39.32 & 41.35 & 40.92 & 42.26 & 40.99 & 50.81 & 49.86 & \textcolor{red}{52.48} & 49.24  &  45.82 & 24.81 & \textcolor{red}{52.39} & \textcolor{blue}{51.09} & 47.12 & 42.41 \\

& LPIPS$\downarrow$ & 0.3575 & 0.1777 & 0.1615 & 0.1710 & 0.2115 & \textcolor{blue}{0.1613} & 0.3635 & 0.1833 & 0.3433 & \textcolor{red}{0.1572} & 0.5571 &  0.3582 & 0.5303 & \textcolor{blue}{0.3238} & 0.5079 & \textcolor{red}{0.3129} \\

& DISTS$\downarrow$ & 0.2005 & 0.1288 & \textcolor{red}{0.1160} & 0.1224 & 0.1436 & 0.1165 & 0.1943 & 0.1372 & 0.1921 & \textcolor{red}{0.1160} & 0.2956 &  0.2096 & 0.3183 & \textcolor{blue}{0.2068} & 0.2753 & \textcolor{red}{0.1972} \\

& NIQE$\downarrow$ & 5.35 & \textcolor{red}{3.18} & 3.53 & \textcolor{blue}{3.23} & 3.67 & 3.23 & 6.83 & 3.51 & 5.10 & 3.30 & 6.23 &  \textcolor{red}{3.15} & 12.82 & \textcolor{blue}{3.68} & 5.49 &  4.58 \\
\hline\hline

General& PSNR$\uparrow$ & \textcolor{red}{30.30} & 27.54 & 27.53 & 27.04 & 28.61 & 27.65 & 29.72 & 27.83 & \textcolor{blue}{29.94} & 28.44  &  \textcolor{blue}{25.38} & 21.51 & 25.09 & 23.45 & \textcolor{red}{25.42} & 24.00 \\

100& SSIM$\uparrow$ & \textcolor{red}{0.8696} & 0.7998 & 0.7984 & 0.7861 & 0.8259 & 0.7995 & 0.8574 & 0.7951 & \textcolor{blue}{0.8629} & 0.8229  &  \textcolor{blue}{0.7081} & 0.5674 & 0.6806 & 0.6063 & \textcolor{red}{0.7097} & 0.6534 \\

& \tiny{LRPSNR}$\uparrow$ & \textcolor{red}{53.96} & 41.44 & 41.93 & 40.05 & 45.06 & 42.31 & 50.65 & 49.59 & \textcolor{blue}{52.22} & 49.82 & 44.78 & 25.19 & \textcolor{red}{48.95} & \textcolor{blue}{47.59} & 44.28 & 41.36 \\

& LPIPS$\downarrow$ & 0.1665 & 0.0962 & 0.0881 & 0.1084 & 0.1118 & \textcolor{blue}{0.0865} & 0.1731 & 0.0962 & 0.1519 & \textcolor{red}{0.0784}& 0.3403 & 0.2494 & 0.3194 & \textcolor{blue}{0.2341} & 0.2924 &  \textcolor{red}{0.2058} \\

& DISTS$\downarrow$ & 0.1321 & 0.0955 & \textcolor{blue}{0.0845} & 0.1166 & 0.1099 & 0.0857 & 0.1276 & 0.1022 & 0.1205 & \textcolor{red}{0.0831}& 0.2362 & \textcolor{blue}{0.1852} & 0.2488 & 0.1899 & 0.2134 & \textcolor{red}{0.1716} \\

& NIQE$\downarrow$ & 6.56 & \textcolor{red}{4.35} & 4.65 & 4.38 & 4.71 & \textcolor{blue}{4.37} & 7.02 & 5.18 & 6.05 & 4.54 & 7.18 & \textcolor{red}{4.40} & 11.92 & \textcolor{blue}{4.89} & 6.09 &  5.46\\
\hline\hline

DIV2K& PSNR$\uparrow$ & \textcolor{red}{29.48} & 26.63 & 26.64 & 26.56 & 27.82 & 26.71 & 29.05 & 27.08 & \textcolor{blue}{29.24} & 27.51  &  \textcolor{blue}{25.50} & 21.37 & 25.09 & 23.04 & \textcolor{red}{25.60} & 23.56 \\

& SSIM$\uparrow$ & \textcolor{red}{0.8444} & 0.7625 & 0.7640 & 0.7578 & 0.7931 & 0.7614 & 0.8290 & 0.7558 & \textcolor{blue}{0.8383} & 0.7890  &  \textcolor{blue}{0.6951} & 0.5533 & 0.6589 & 0.5728 & \textcolor{red}{0.6989} & 0.6241 \\

& \tiny{LRPSNR}$\uparrow$ & \textcolor{red}{53.72} & 40.87 & 42.61 & 40.40 & 44.64 & 42.57 & 51.02 & 49.96 & \textcolor{blue}{53.30} & 50.54 & 46.05 & 25.21 & \textcolor{red}{51.28} & \textcolor{blue}{50.26} & 46.96 & 42.66
 \\

& LPIPS$\downarrow$ & 0.2537 & 0.1263 & 0.1154 & 0.1449 & 0.1523 & \textcolor{blue}{0.1099} & 0.2513 & 0.1201 & 0.2390 & \textcolor{red}{0.1028}& 0.4245 & 0.2841 & 0.4033 & \textcolor{blue}{0.2719} & 0.3857 & \textcolor{red}{0.2403} \\

& DISTS$\downarrow$ & 0.1261 & 0.0613 & 0.0530 & 0.0858 & 0.0766 & \textcolor{red}{0.0493} & 0.1139 & 0.0622 & 0.1169 & \textcolor{blue}{0.0513}& 0.2203 & \textcolor{blue}{0.1293} & 0.2342 & 0.1386 & 0.1953 & \textcolor{red}{0.1190} \\

& NIQE$\downarrow$ & 4.42 & \textcolor{red}{2.57} & 2.79 & 2.92 & 2.91 & \textcolor{blue}{2.74} & 5.16 & 3.50 & 4.11 & 2.81& 5.15 & \textcolor{red}{2.53} & 7.15 & \textcolor{blue}{3.54} & 4.41 & 3.61 \\
\hline\hline
Param.          & {} & 16.7M  & 1.5M   & 16.7M  & 53.7M  & 4.8M   & 24.8M  & 39.5M            &  39.5M  & 18.3M     & 18.3M           & 16.7M & 16.7M & 50.8M & 50.8M & 18.3M & 18.3M \\ 
\hline
\end{tabularx}
%\end{table}
%\end{tabularx}
% \end{tabular}
\end{center}
\label{tab:tab01}
\end{table*}
\subsubsection{Evaluation Method} 
To evaluate the perceptual distance to the Ground Truth, we report LPIPS ~\cite{zhang2018unreasonable} as default ~\cite{lugmayr2019unsupervise}, and additionally use DISTS ~\cite{ding2020image} as structure and texture similarity in some cases. PSNR and SSIM ~\cite{wang2004image} are reported as fidelity-oriented metrics. Furthermore, we report the no-reference metric NIQE ~\cite{zhang2018unreasonable}. Since the consistency with the LR image is also an important factor, we report the LR-PSNR, computed as the PSNR between the downsampled SR image and the original LR. To measure the meaningful diversity of SR methods that can actively sample from the space of plausible super-resolutions, we also report the SR-Diversity score, which is used for the evaluation protocol on the \href{https://github.com/andreas128/NTIRE21_Learning_SR_Space}{Super-Resolution Space Challenge learning track in the NTIRE Challenge 2021 ~\cite{SRSpace2021, Lugmayr_2021_CVPR}.} Specifically, we sample 11 images and densely calculate LPIPS ~\cite{zhang2018unreasonable} metric between the samples and the ground truth. To obtain the local best score, we pixel-wisely select the best score out of the 11 samples and take the full image's average. The global best score is calculated by averaging the whole image's score and selecting the best. Then, the diversity score is calculated as follows:
\begin{equation}
score = (global best - local best)/(global best) \times 100.
\label{eqn:div_score}
\end{equation}

%%%%%%%% figure PD 4x %%%%%%%%
\begin{figure*}[!t]

\setlength{\arrayrulewidth}{1.0pt}
\newcolumntype{Z}
{>{\centering\arraybackslash}X}
\begin{center}
\small
%\footnotesize
% \scriptsize
\renewcommand{\tabcolsep}{1pt}
\begin{tabularx}{\linewidth}{Z Z Z Z Z Z Z Z Z }
    \multicolumn{9}{c}{{4}$\times$ SR comparison}\\
\hline
      Whole image  & HR & RRDB~\cite{2018esrgan} & ESRGAN~\cite{2018esrgan} & SFTGAN~\cite{2018recovering} & NatSR~\cite{2019natural} & SPSR~\cite{2020structure} & SRFlow $t$=0.9 & FxSR $t$=0.8\\
\hline
\end{tabularx}
\end{center}

\centering
% \scriptsize
\begin{minipage}[t]{1.0\linewidth}
    \centering
    
    \includegraphics[width=0.105\linewidth]{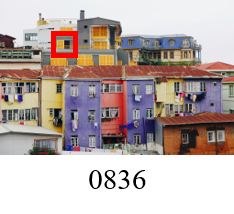}\hfill
    \includegraphics[width=0.105\linewidth]{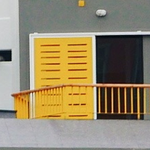}\hfill
    \includegraphics[width=0.105\linewidth]{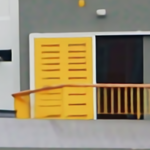}\hfill
    \includegraphics[width=0.105\linewidth]{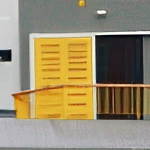}\hfill
    \includegraphics[width=0.105\linewidth]{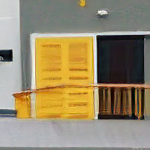}\hfill
    \includegraphics[width=0.105\linewidth]{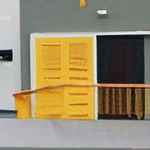}\hfill
    \includegraphics[width=0.105\linewidth]{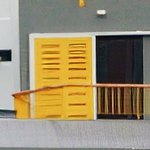}\hfill
    \includegraphics[width=0.105\linewidth]{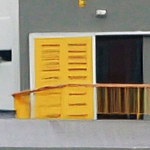}\hfill
    \includegraphics[width=0.105\linewidth]{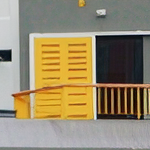}\vfill
    \vspace{0.4cm}
    
    \includegraphics[width=0.105\linewidth]{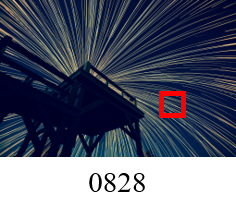}\hfill
    \includegraphics[width=0.105\linewidth]{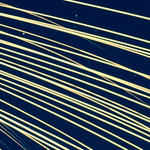}\hfill
    \includegraphics[width=0.105\linewidth]{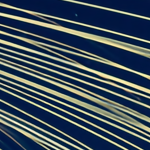}\hfill
    \includegraphics[width=0.105\linewidth]{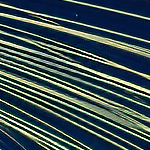}\hfill
    \includegraphics[width=0.105\linewidth]{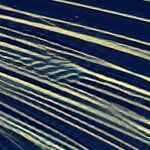}\hfill
    \includegraphics[width=0.105\linewidth]{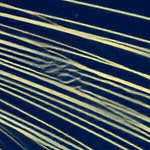}\hfill
    \includegraphics[width=0.105\linewidth]{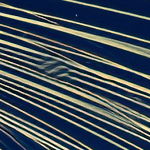}\hfill
    \includegraphics[width=0.105\linewidth]{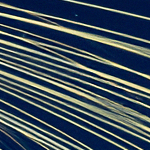}\hfill
    \includegraphics[width=0.105\linewidth]{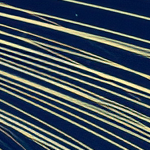}\vfill
    \vspace{0.4cm}
    
    \includegraphics[width=0.105\linewidth]{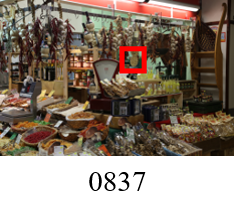}\hfill
    \includegraphics[width=0.105\linewidth]{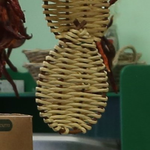}\hfill
    \includegraphics[width=0.105\linewidth]{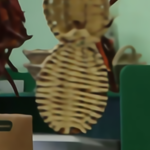}\hfill
    \includegraphics[width=0.105\linewidth]{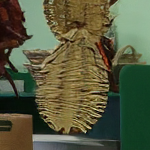}\hfill
    \includegraphics[width=0.105\linewidth]{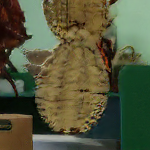}\hfill
    \includegraphics[width=0.105\linewidth]{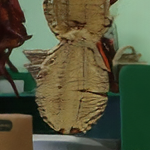}\hfill
    \includegraphics[width=0.105\linewidth]{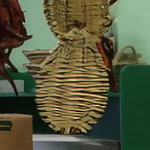}\hfill
    \includegraphics[width=0.105\linewidth]{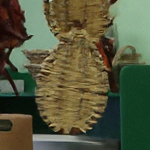}\hfill
    \includegraphics[width=0.105\linewidth]{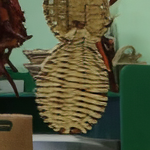}\vfill
    \vspace{0.4cm}
    
    \includegraphics[width=0.105\linewidth]{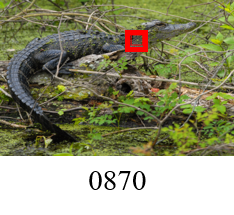}\hfill
    \includegraphics[width=0.105\linewidth]{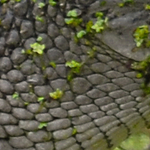}\hfill
    \includegraphics[width=0.105\linewidth]{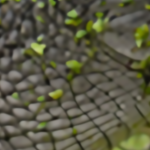}\hfill
    \includegraphics[width=0.105\linewidth]{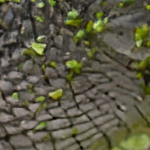}\hfill
    \includegraphics[width=0.105\linewidth]{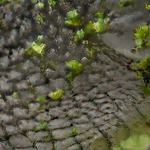}\hfill
    \includegraphics[width=0.105\linewidth]{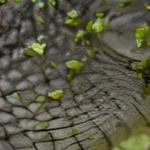}\hfill
    \includegraphics[width=0.105\linewidth]{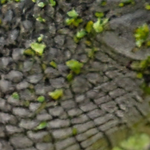}\hfill
    \includegraphics[width=0.105\linewidth]{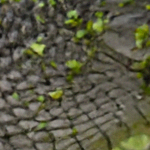}\hfill
    \includegraphics[width=0.105\linewidth]{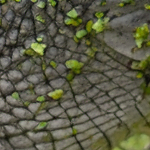}\vfill
    \vspace{0.1cm}
    
    \includegraphics[width=0.105\linewidth]{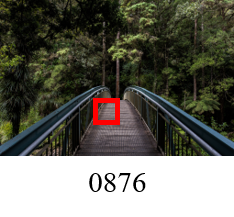}\hfill
    \includegraphics[width=0.105\linewidth]{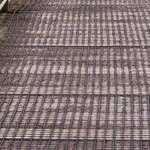}\hfill
    \includegraphics[width=0.105\linewidth]{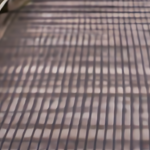}\hfill
    \includegraphics[width=0.105\linewidth]{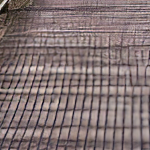}\hfill
    \includegraphics[width=0.105\linewidth]{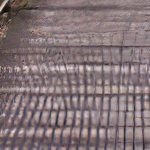}\hfill
    \includegraphics[width=0.105\linewidth]{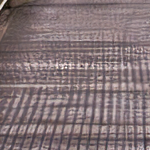}\hfill
    \includegraphics[width=0.105\linewidth]{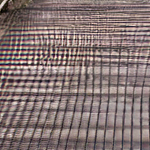}\hfill
    \includegraphics[width=0.105\linewidth]{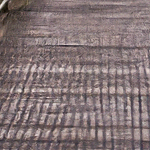}\hfill
    \includegraphics[width=0.105\linewidth]{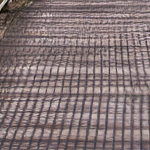}\vfill

\end{minipage}
\caption{Visual comparison with state-of-the-art perception-driven SR methods on DIV2K validation set~\cite{agustsson2017ntire}. The proposed method produces competitive results compared to other modern techniques and can also generate reconstructed images of various styles of LR images.}
\label{fig:fig_comp_4x}
\end{figure*}

%%%%%%%% fig_FxSR-PD comp 8x %%%%%%%% 
\begin{figure*}[!t]

\setlength{\arrayrulewidth}{1.0pt}
\newcolumntype{Z}
{>{\centering\arraybackslash}X}
\begin{center}
\small
\renewcommand{\tabcolsep}{1pt}
\begin{tabularx}{\linewidth}{Z Z Z Z Z Z }
% \hline
    \multicolumn{6}{c}{{8}$\times$ SR comparison}\\
\hline
      Whole image  & HR & RRDB~\cite{2018esrgan} & ESRGAN~\cite{2018esrgan} & SRFlow $t$=0.9 & FxSR $t$=0.8\\
\hline
\end{tabularx}
\end{center}

\begin{minipage}[t]{1.0\linewidth}
    \centering
    \includegraphics[width=0.14\linewidth]{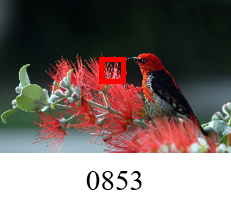}\hfill
   \includegraphics[width=0.14\linewidth]{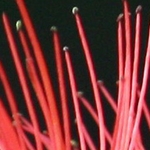}\hfill
    \includegraphics[width=0.14\linewidth]{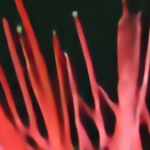}\hfill
    \includegraphics[width=0.14\linewidth]{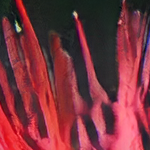}\hfill
    \includegraphics[width=0.14\linewidth]{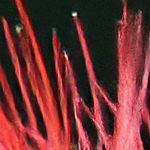}\hfill
    \includegraphics[width=0.14\linewidth]{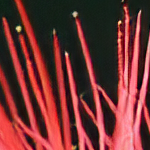}\vfill

    \centering
    \includegraphics[width=0.14\linewidth]{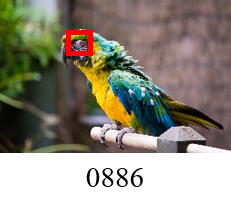}\hfill
    \includegraphics[width=0.14\linewidth]{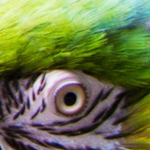}\hfill
    \includegraphics[width=0.14\linewidth]{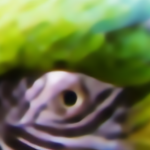}\hfill
    \includegraphics[width=0.14\linewidth]{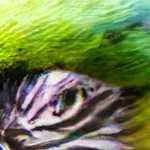}\hfill
    \includegraphics[width=0.14\linewidth]{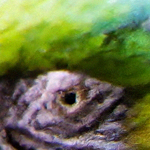}\hfill
    \includegraphics[width=0.14\linewidth]{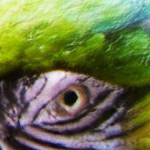}\vfill
\end{minipage}
\caption{Visual comparison for 8$\times$ SR results on DIV2K validation set~\cite{agustsson2017ntire}.}
\label{fig:fig_comp_8x}
\end{figure*}

\subsubsection{Training Method}
For the scaling factor ${4}\times$, sub-images are cropped with the sizes of ${320}\times{320}$ with a stride of $160$ and ${80}\times{80}$ with ${40}$, for the HR and LR training images, respectively. For the scaling factor ${8}\times$, the LR sub-images are cropped to the size of ${40}\times{40}$ with a ${20}$ strides. Then, the batch image pairs for each iteration of training are randomly cropped from these sub-images. The HR batch size is ${128}\times{128}$ and the LR batch sizes are ${32}\times{32}$ and ${16}\times{16}$ for scaling factors of ${4}\times$ and ${8}\times$, respectively. 

For the optimization, we use initial learning rate of ${10}^{-4}$. The learning rate is halved after 5K, 10K, 20K, and 30K iterations. Adam~\cite{KingmaB14} with ${\beta}_{1}=0.9$ and ${\beta}_{2}=0.99$ is used for both generator and discriminator training. We use pre-trained RRDB ~\cite{2018esrgan} and ESRGAN ~\cite{2018esrgan} models to optimize the proposed FxSR models. While fine-tuning FxSR-PD and FxSR-DS, ${\lambda}_{rec\_o}$, ${\lambda}_{adv\_o}$ and ${\lambda}_{per}$ are set to be ${1}\times{10}^{-2}$, ${5}\times{10}^{-3}$ and ${1.0}$ respectively, but $\eta$ is set differently to ${1}\times{10}$ and ${1.0}$.

% \hl{The quantitative and visual comparison with state-of-the-art SR methods are provided. We also report the performance in terms of diversity and the LR consistency of the SR results as well as the complexity of the proposed model. Then, we demonstrate some examples of applying local style control and show validation on compressed LR images.}

%%%%%%%% fig_FxSR-DS T %%%%%%%% %%%%%%%% 
\begin{figure*}[!t]

\setlength{\arrayrulewidth}{1.0pt}
\newcolumntype{Z}
{>{\centering\arraybackslash}X}
\begin{center}
\small
%\footnotesize

\renewcommand{\tabcolsep}{1pt}
\begin{tabularx}{\linewidth}{Z Z Z Z Z Z}
\hline
      Whole image  & HR & \multicolumn{4}{c}{$4\times$ FxSR-DS} \\
\hline  \\
&&\multicolumn{4}{c}{\includegraphics[width=0.55\linewidth]{figure/control_bar.png}}\\
        &  & $t=0.0$ & $t=0.3$ & $t=0.6$ & $t=1.0$ \\
\end{tabularx}
\end{center}

\centering
% \scriptsize
\begin{minipage}[t]{1.0\linewidth}
    \centering
    
    \includegraphics[width=0.16\linewidth]{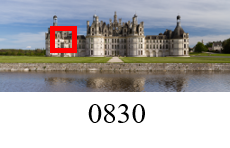}\hfill
    \includegraphics[width=0.16\linewidth]{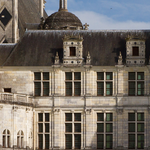}\hfill
    \includegraphics[width=0.16\linewidth]{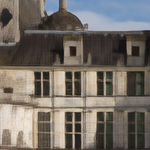}\hfill
    \includegraphics[width=0.16\linewidth]{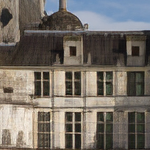}\hfill
    \includegraphics[width=0.16\linewidth]{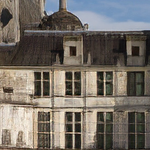}\hfill
    \includegraphics[width=0.16\linewidth]{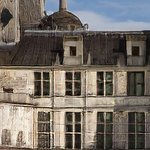}\vfill
    \vspace{0.3cm}
    
    \includegraphics[width=0.16\linewidth]{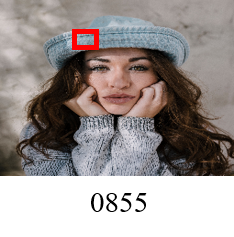}\hfill
    \includegraphics[width=0.16\linewidth]{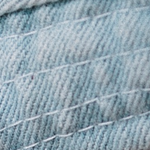}\hfill
    \includegraphics[width=0.16\linewidth]{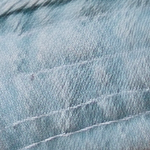}\hfill
    \includegraphics[width=0.16\linewidth]{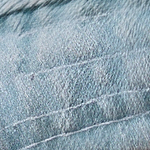}\hfill
    \includegraphics[width=0.16\linewidth]{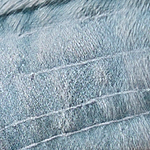}\hfill
    \includegraphics[width=0.16\linewidth]{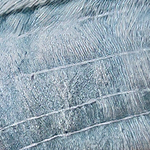}\vfill
    \vspace{0.3cm}
    
    \includegraphics[width=0.16\linewidth]{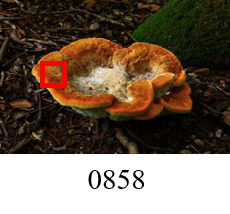}\hfill
    \includegraphics[width=0.16\linewidth]{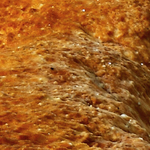}\hfill
    \includegraphics[width=0.16\linewidth]{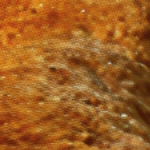}\hfill
    \includegraphics[width=0.16\linewidth]{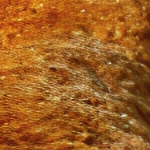}\hfill
    \includegraphics[width=0.16\linewidth]{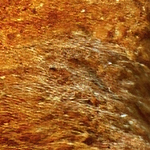}\hfill
    \includegraphics[width=0.16\linewidth]{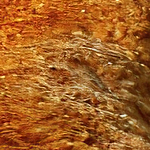}\vfill
    \vspace{0.3cm}

    % \includegraphics[width=0.16\linewidth]{figure/fig_FxSR-DS_0864_00.png}\hfill
    % \includegraphics[width=0.16\linewidth]{figure/fig_FxSR-DS_0864_01.png}\hfill
    % \includegraphics[width=0.16\linewidth]{figure/fig_FxSR-DS_0864_02.png}\hfill
    % \includegraphics[width=0.16\linewidth]{figure/fig_FxSR-DS_0864_03.png}\hfill
    % \includegraphics[width=0.16\linewidth]{figure/fig_FxSR-DS_0864_04.png}\hfill
    % \includegraphics[width=0.16\linewidth]{figure/fig_FxSR-DS_0864_05.png}\vfill
    % \vspace{0.3cm}
    
    \includegraphics[width=0.16\linewidth]{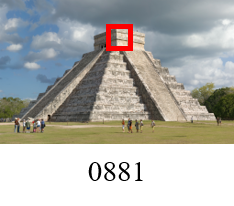}\hfill
    \includegraphics[width=0.16\linewidth]{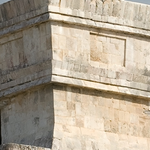}\hfill
    \includegraphics[width=0.16\linewidth]{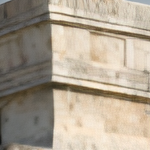}\hfill
    \includegraphics[width=0.16\linewidth]{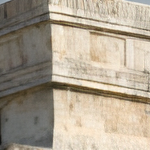}\hfill
    \includegraphics[width=0.16\linewidth]{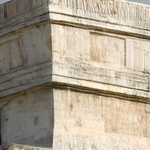}\hfill
    \includegraphics[width=0.16\linewidth]{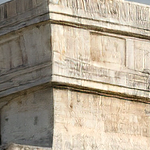}\vfill
    \vspace{0.3cm}

    % \includegraphics[width=0.16\linewidth]{figure/fig_FxSR-DS_0883_00.png}\hfill
    % \includegraphics[width=0.16\linewidth]{figure/fig_FxSR-DS_0883_01.png}\hfill
    % \includegraphics[width=0.16\linewidth]{figure/fig_FxSR-DS_0883_02.png}\hfill
    % \includegraphics[width=0.16\linewidth]{figure/fig_FxSR-DS_0883_03.png}\hfill
    % \includegraphics[width=0.16\linewidth]{figure/fig_FxSR-DS_0883_04.png}\hfill
    % \includegraphics[width=0.16\linewidth]{figure/fig_FxSR-DS_0883_05.png}\vfill
    % \vspace{0.3cm}
    
    \includegraphics[width=0.16\linewidth]{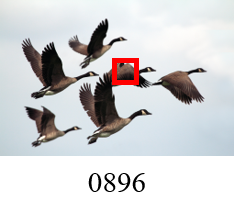}\hfill
    \includegraphics[width=0.16\linewidth]{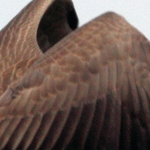}\hfill
    \includegraphics[width=0.16\linewidth]{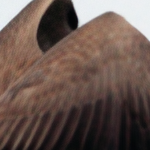}\hfill
    \includegraphics[width=0.16\linewidth]{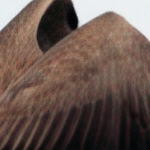}\hfill
    \includegraphics[width=0.16\linewidth]{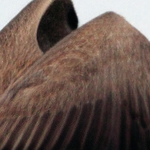}\hfill
    \includegraphics[width=0.16\linewidth]{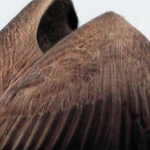}\vfill
\end{minipage}
\caption{Changes in the result of FxSR-DS $4\times$ SR according to $t$ on DIV2K validation set~\cite{agustsson2017ntire} .}
\label{fig:fig_FxSR-DS_t}
\end{figure*}

%%%%%%%% fig_FxSR-DS T %%%%%%%% %%%%%%%% 
\begin{figure*}[!t]
\centering
% \scriptsize
\begin{minipage}[t]{1.0\linewidth}
    \centering
    \subfigure[The FxSR-PD result with G-Best LPIPS] {\includegraphics[width=0.3\linewidth]{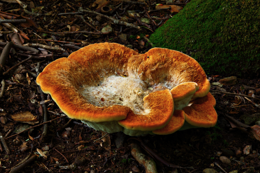}}\hfill
    \subfigure[The G-Best LPIPS map (0.1355)] {\includegraphics[width=0.3\linewidth]{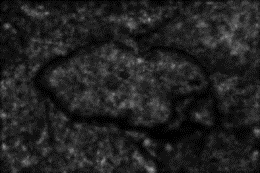}}\hfill
    \subfigure[The L-Best LPIPS map (0.1289)] {\includegraphics[width=0.3\linewidth]{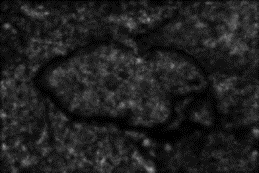}}\vfill
    \vspace{0.3cm}

    \subfigure[The FxSR-DS result with G-Best LPIPS] {\includegraphics[width=0.3\linewidth]{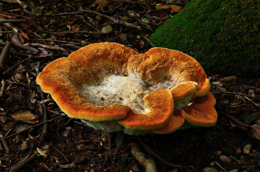}}\hfill
    \subfigure[The G-Best LPIPS map (0.1404)] {\includegraphics[width=0.3\linewidth]{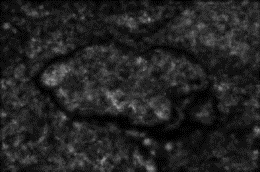}}\hfill
    \subfigure[The L-Best LPIPS map (0.1168)] {\includegraphics[width=0.3 \linewidth]{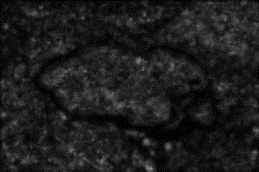}}\vfill
    \vspace{0.3cm}
    
\end{minipage}
\caption{On the left are the SR results of FxSR-PD (top) and FxSR-DS (bottom) for DIV2K 0858, corresponding to t values with Global Best (G-Best) LPIPIS among 11 samples, respectively. In the middle are the LPIPS maps of the SR results on the left. On the right are the Local Best (L-Best) LPIPS maps generated by selecting the highest score per pixel from 11 samples. The brighter the pixel, the higher the LPIPS value and the greater the perceptual difference from the ground truth. Each number in parentheses is the average LPIPS value for the entire image.}
\label{fig:fig_DS_DIV_LOC}
\end{figure*}

% \subsection{Results and Analysis}
\subsection{Evaluation of Flexible SR for Perception-Distortion (FxSR-PD)}
% \hl{\mbox{Figure~\ref{fig:fig_FxSR-PD_t}} shows that the FxSR-PD model can produce SR results with different trade-off between distortion and perception objectives by adjusting a single parameter at the inference phase.}
By adjusting a single parameter $t$, the FxSR-PD model can generate various SR results for the trade-offs between distortion and perception objective at the inference phase, as shown in Figure ~\ref{fig:fig_FxSR-PD_t}. It shows that $t=0$ generates blurry outputs as the FxSR objective is distortion-oriented, and $t=1$ generates sharp textures as the FxSR becomes perception-oriented. Also, the $t$ between 0 and 1 generates different trade-offs, with less or more distortions, and more or less blurriness.
\subsubsection{Quantitative Comparison}
We compare our method quantitatively with distortion-oriented methods such as RRDB ~\cite{2018esrgan}, and perception-oriented methods such as SRGAN ~\cite{2017photo}, ESRGAN ~\cite{2018esrgan}, SFTGAN ~\cite{2018recovering}, NatSR ~\cite{2019natural}, SPSR ~\cite{2020structure} and SRFlow ~\cite{2020srflow}. For the $4\times$ SR, we use pre-trained models provided by the authors, while for the non-provided $8\times$ SR, we used the author's code to train the RRDB ~\cite{2018esrgan} and ESRGAN ~\cite{2018esrgan} models. The results are presented in Figures from ~\ref{fig:fig_PSNR_4x} to ~\ref{fig:fig_P-D_4x} and Table ~\ref{tab:tab01}.
Figures from ~\ref{fig:fig_PSNR_4x} to ~\ref{fig:fig_NIQE_4x} show the performance comparison of $4\times$ SR results according to $t$, evaluated by the distortion-oriented (PSNR, SSIM ~\cite{wang2004image}, MS-SSIM ~\cite{wang2003multiscale}), perception-oriented (LPIPS ~\cite{zhang2018unreasonable}, DISTS ~\cite{ding2020image}, VIF ~\cite{sheikh2006image}), and non-reference perception-oriented metrics (NIQE ~\cite{mittal2012making}, BRISQUE ~\cite{mittal2012no}), respectively. In Figure ~\ref{fig:fig_PSNR_4x}, we can see that the scores of the distortion-oriented metrics improve as $t$ approaches 0, whereas in Figures ~\ref{fig:fig_LPIPS_4x} and ~\ref{fig:fig_NIQE_4x}, the scores of the perception-oriented metrics improve as $t$ approaches 1.

Since there is a trade-off between the distortion-oriented metrics and the perception-oriented metrics, it is necessary to evaluate the performance of the SR models in a perception-distortion 2D plane ~\cite{blau2018perception}, as shown in Figure~\ref{fig:f01}. The vertical axis denotes perceptual loss LPIPS~\cite{zhang2018unreasonable}, and the horizontal axis the PSNR (distortion-oriented measure). Hence, the lower left part is the desired place where both MSE and perceptual loss are low ~\cite{blau2018perception}, and we can see that our method is comparable to others in this respect. Note that the RRDB ~\cite{2018esrgan} and ESRGAN ~\cite{2018esrgan} are the results of using distortion-oriented and perception-oriented loss, respectively. Others drawn in solid lines are adjustable methods. Pixel interpolation (Pix-Interp) and network weight interpolation (Net-Interp) methods utilize two differently trained models, {\em i.e.}, the RRDB and ESRGAN stated above. The number of parameters for each method is also provided for complexity comparison. More details about complexity analysis will be provided in Section IV.F.

Since various metrics examined in Figures ~\ref{fig:fig_PSNR_4x}-~\ref{fig:fig_NIQE_4x} have different characteristics and performance, we present additional performance comparisons for the perception-distortion plane with these metrics in Figure ~\ref{fig:fig_P-D_4x}. These comparisons show trends similar to those in Figure ~\ref{fig:f01}.
Table ~\ref{tab:tab01} shows the evaluation of FxSR-PD and other SR methods for the specific $t$ values. The proposed FxSR-PD obtains the best PSNR and SSIM at $t=0$ among perception-oriented methods and the best LPIPS values at $t=0.8$ for all datasets.

%%%%%%%% Table 2 Diversity %%%%%%%% %%%%%%%% %%%%%%%% 
% {\renewcommand{\arraystretch}{1.02}
\begin{table}[!ht]
\caption{Comparison with state-of-the-art SR methods on DIV2K in terms of low resolution consistency, photo-realism and meaningful diversity. The numbers in the parentheses are the relative performances, i.e., the best value is set to 1, and the others are divided by the best value.}
% \vspace{-0.3cm}
% \newcolumntype{Z}
% {>{\centering\arraybackslash}X}

\begin{center}
% \small
%\footnotesize
\scriptsize
\centering
\begin{tabular}{|c|c|c|c|c|c|c|}
\hline
      {}  & SR  & LR- & Mean  & G-best  & L-best & Div. \\
      {}  & Model  & PSNR$\uparrow$ & LPIPS$\downarrow$  &  LPIPS$\downarrow$ & LPIPS$\downarrow$  &score$\uparrow$ \\
\hline\hline
% \vspace{0.3cm}
\multirow{8}*{4$\times$} & \multirow{2}* {SRFlow~\cite{2020srflow}} & 50.55 & 0.1765 & 0.1153 & 0.0905 &\textcolor{red}{23.12}      \\
 ~&  & (0.99) & (1.54) & (1.14) & (1.03) & \textcolor{red}{(1.00)} \\
\cline{2-7}
~&\multirow{2}* {DNI~\cite{wang2019deep}} & 44.37 & 0.1968 & 0.1114 & 0.1003 & 10.01 \\
 ~&  & (0.87) & (1.72) & (1.10) & (1.14)	& (0.43) \\
\cline{2-7}
~&\multirow{2}* {FxSR-PD} & \textcolor{red}{51.16} & 0.1253 & \textcolor{red}{0.1010} & 0.0926 & 8.98  \\
 ~&  & \textcolor{red}{(1.00)} & (1.10) & \textcolor{red}{(1.00)} & (1.05) & (0.39) \\
\cline{2-7}
~&\multirow{2}* {FxSR-DS} & 44.49 & \textcolor{red}{0.1144} & 0.1018 & \textcolor{red}{0.0880} & 13.66 \\
 ~&  & (0.87) & \textcolor{red}{(1.00)} & (1.01) & (1.00) & (0.59) \\
\hline\hline
\multirow{6}*{8$\times$} & \multirow{2}* {SRFlow~\cite{2020srflow}} & \textcolor{red}{50.78} & 0.3261 & 0.2613 & 0.2066 & \textcolor{red}{21.88} \\
 ~&  & (1.00) & (1.32) & (1.19) & (1.08)	& \textcolor{red}{(1.00)} \\
\cline{2-7}
~&\multirow{2}* {FxSR-PD} & 44.76 & \textcolor{red}{0.2477} & \textcolor{red}{0.2192} & 0.1996 & 9.11     \\
 ~&  & (0.88)	& \textcolor{red}{(1.00)}	& \textcolor{red}{(1.00)}	& (1.04)	& (0.42) \\
\cline{2-7}
~&\multirow{2}* {FxSR-DS} & 37.77 & \textcolor{red}{0.2477} & 0.2206 & \textcolor{red}{0.1912} & 13.39     \\
 ~&  & (0.74) & \textcolor{red}{(1.00)}	& (1.01)	& \textcolor{red}{(1.00)}	& (0.61) \\
% \hline
%       {}  & SR  & LR- & Mean  & G-best  & L-best & Div. \\
%       {}  & Model  & PSNR$\uparrow$ & LPIPS$\downarrow$  &  LPIPS$\downarrow$ & LPIPS$\downarrow$  &score$\uparrow$ \\
% \hline\hline
% {4$\times$} & SRFlow~\cite{2020srflow} & 49.96 & 0.1811 & 0.1158 & 0.0976  &\textcolor{red}{17.57}      \\
% &DNI~\cite{wang2019deep} & 44.54 & 0.2050 & 0.1232 & 0.1176 & 4.93      \\
% &FxSR-PD & \textcolor{red}{51.29} & 0.1274 & \textcolor{red}{0.1013} & 0.0937 & 8.14         \\
% &FxSR-DS & 44.56 & \textcolor{red}{0.1149} & 0.1018 & \textcolor{red}{0.0892} & 12.46      \\
% \hline\hline
% {8$\times$} & SRFlow~\cite{2020srflow} & \textcolor{red}{51.08} & 0.3316 & 0.2662 & 0.2198 & \textcolor{red}{18.30}      \\
% &FxSR-PD & 44.90 & 0.2501 & \textcolor{red}{0.2193} & 0.2010 & 8.60     \\
% &FxSR-DS & 37.83 & \textcolor{red}{0.2494} & 0.2207 & \textcolor{red}{0.1940} & 12.18     \\

% {4$\times$} & SRFlow~\cite{2020srflow} & 49.96 & 0.1765 & 0.1765 & 0.1765 & \textcolor{red}{23.1}      \\
% & DNI~\cite{wang2019deep} & 41.59 & 0.2050 & 0.1232 & 0.1176 & 4.93      \\
% &FxSR-PD & \textcolor{red}{50.08} & 0.1274 & 8.1        \\
% &FxSR-DS & 43.77 & \textcolor{red}{0.1149} & 12.5      \\
% \hline\hline
% {8$\times$} & SRFlow~\cite{2020srflow} & 0.3261 & 50.00 & 21.8      \\
% &FxSR-PD & 0.2558 & 43.53 & 10.2     \\
% &FxSR-PD & 0.2558 & 36.53 & 10.2     \\
\hline
\end{tabular}
\end{center}
\label{tab:tab_div}
\end{table}
% }

%%%%%%%% figure Local Map %%%%%%%% %%%%%%%% %%%%%%%% 
% \begin{figure}[!t]
\begin{figure*}[!t]
\centering
\scriptsize
\begin{minipage}[t]{1.0\linewidth}
    \centering
    \subfigure[The conventional method of using multiple SR models trained separately for a different objective each.] {\includegraphics[width=0.82\linewidth
    ]{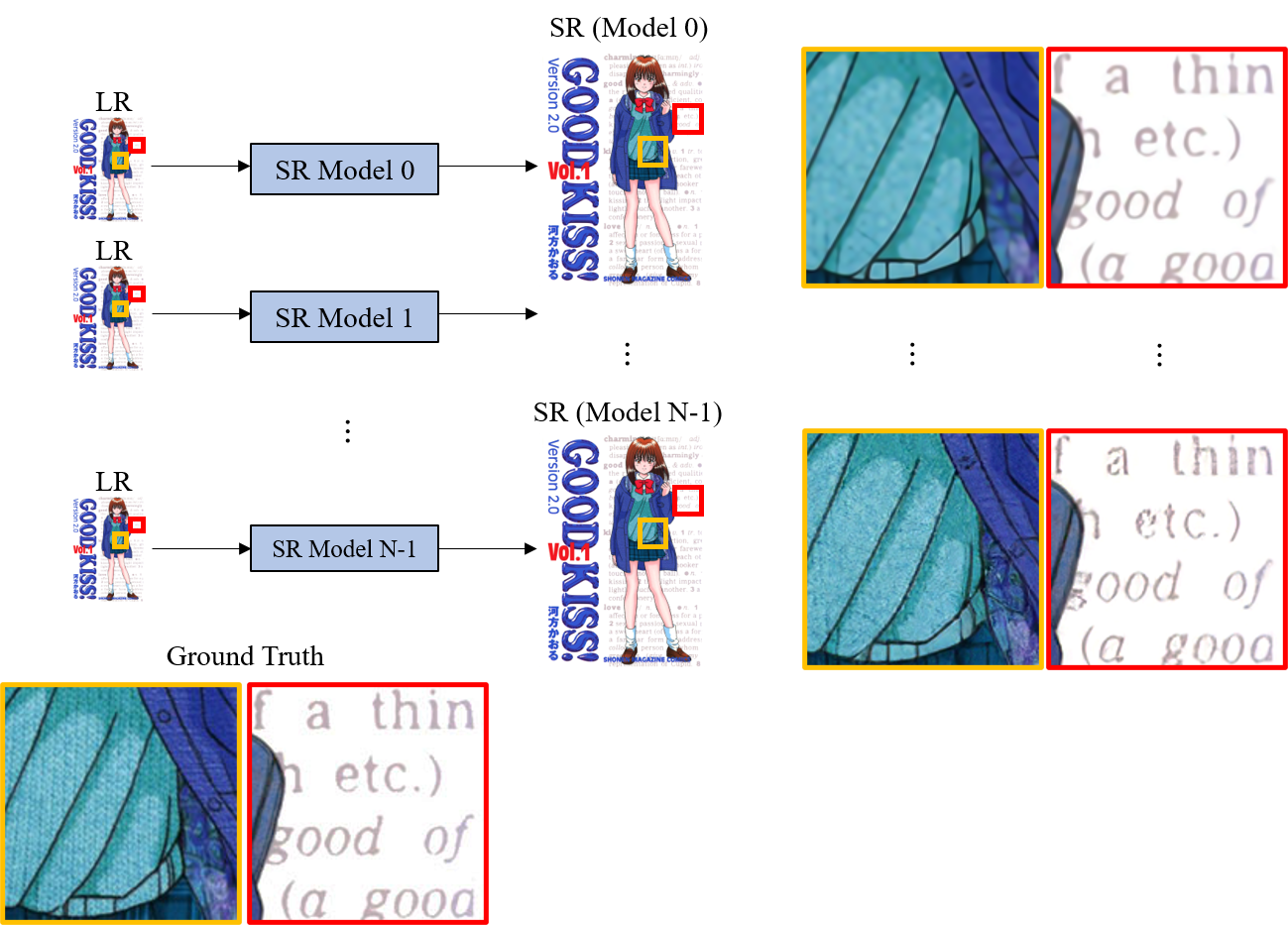}}\vfill
    \subfigure[The proposed method of using single FxSR-PD model trained on the training distribution of objectives.] {\includegraphics[width=0.82\linewidth
    ]{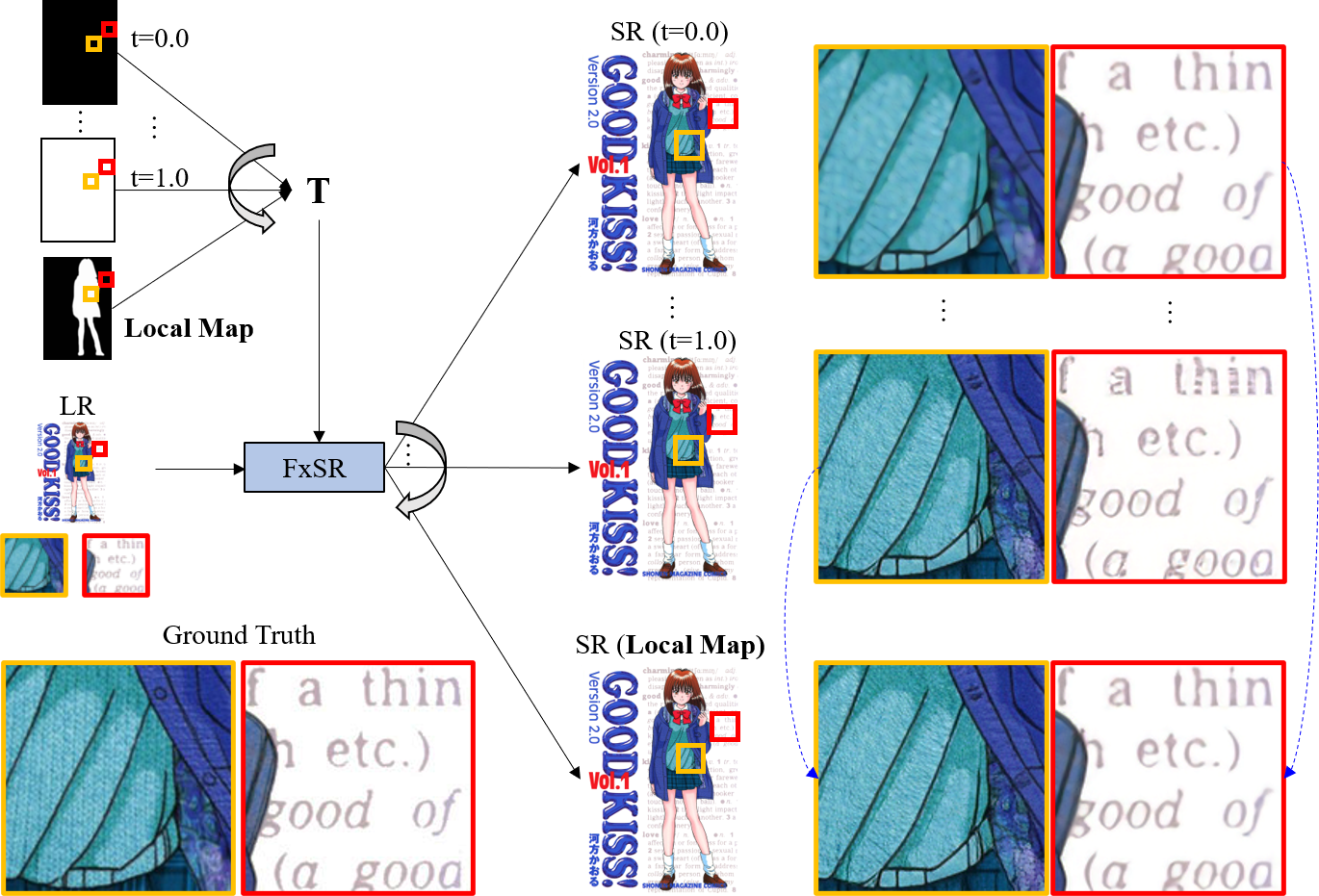}}\vfill
\end{minipage}
\caption{Comparison of the SR results of the conventional method (a), which applies one objective to the entire image, and the FxSR-PD method, which applies different objectives for each area (clothes and letters) through a local map. We can see that the proposed FxSR-PD in (b) can more accurately produce the locally intended and suitable SR results without side effects such as blurry textures and broken characters.}
\label{fig:fig_Local_Map1}
\end{figure*}
%%%%%%%% figure Local Map %%%%%%%% %%%%%%%% %%%%%%%% 
% \begin{figure}[!t]
\begin{figure*}[!t]
\centering
\scriptsize
\begin{minipage}[t]{1.0\linewidth}
    \centering
    \subfigure[The conventional method of using multiple SR models trained separately for a different objective each.] {\includegraphics[width=0.82\linewidth
    ]{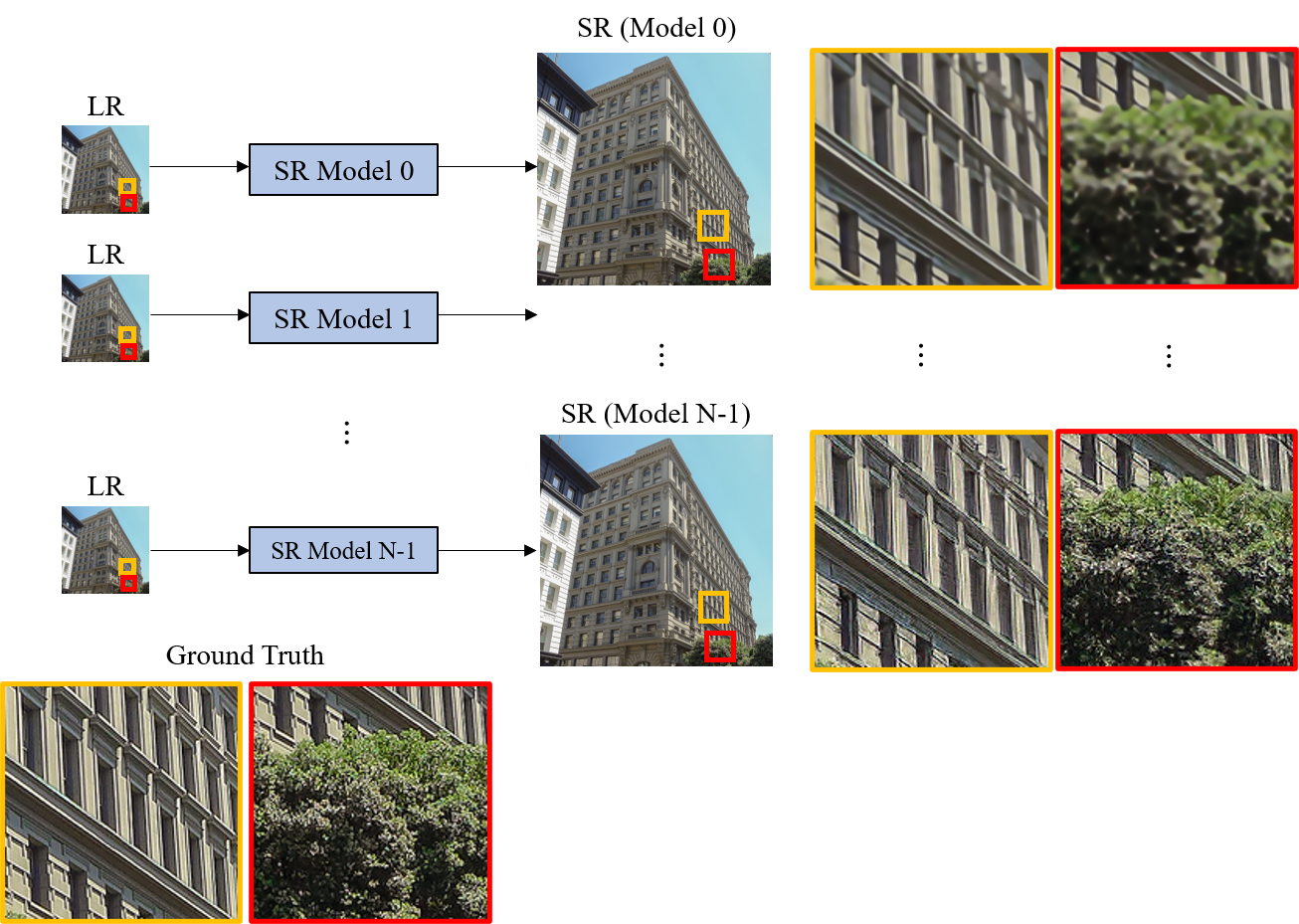}}\vfill
    \subfigure[The proposed method of using single FxSR-DS model trained on the training distribution of objectives.] {\includegraphics[width=0.82\linewidth
    ]{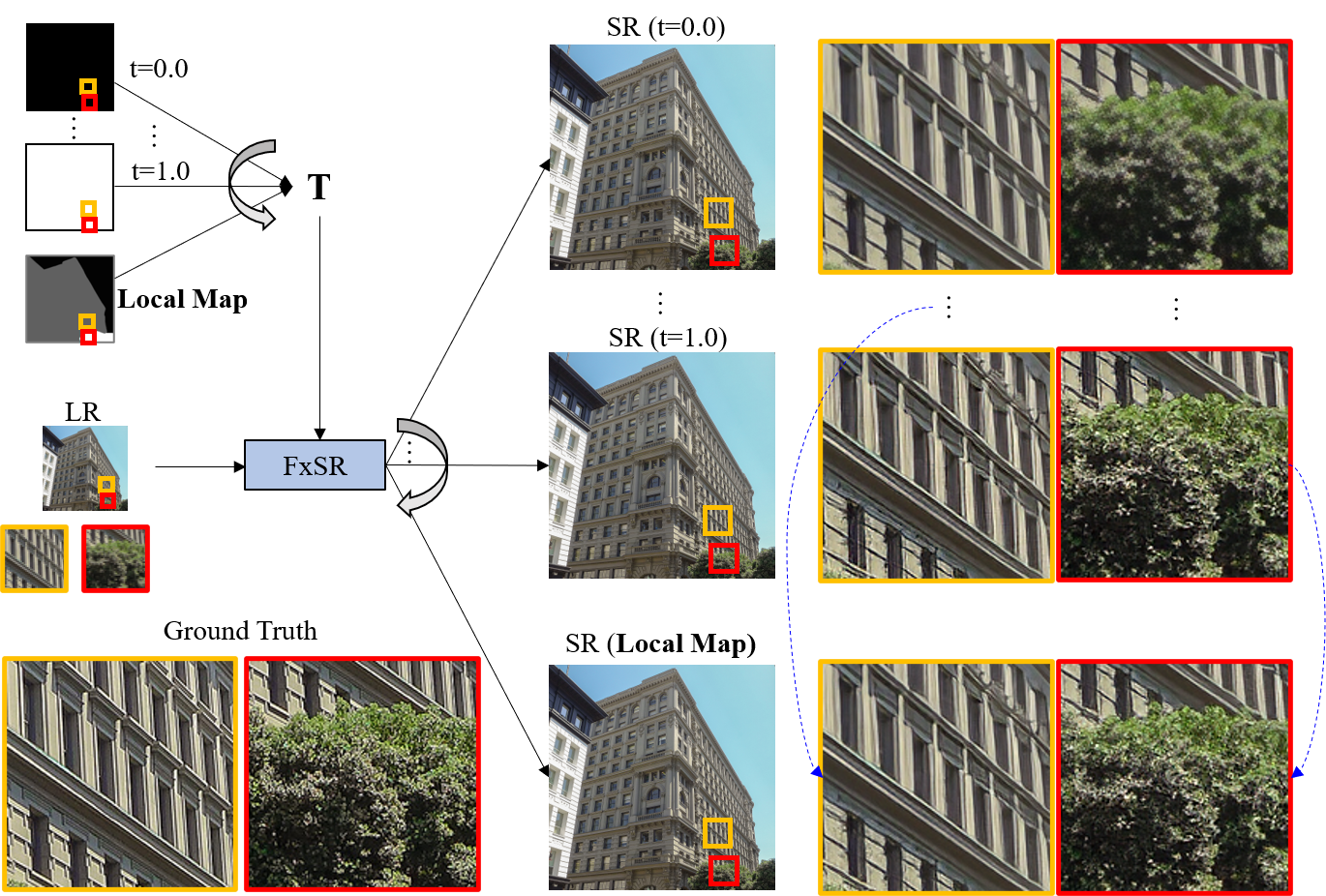}}\vfill
\end{minipage}
\caption{Comparison of the SR results of the conventional method (a), which applies one objective to the entire image, and the FxSR-DS method, which applies different objectives for each area (buildings and trees) through a local map. We can see that the proposed FxSR-DS in (b) can more accurately produce the locally intended and suitable SR results without side effects such as blurry tree textures and overshoot around the edges.}
\label{fig:fig_Local_Map2}
\end{figure*}

%%%%%%%% figure Depth %%%%%%%% %%%%%%%% %%%%%%%% 
\begin{figure*}[!t]
\centering
\scriptsize
\begin{minipage}[t]{1.0\linewidth}
    \centering
    {\includegraphics[width=0.85\linewidth
    ]{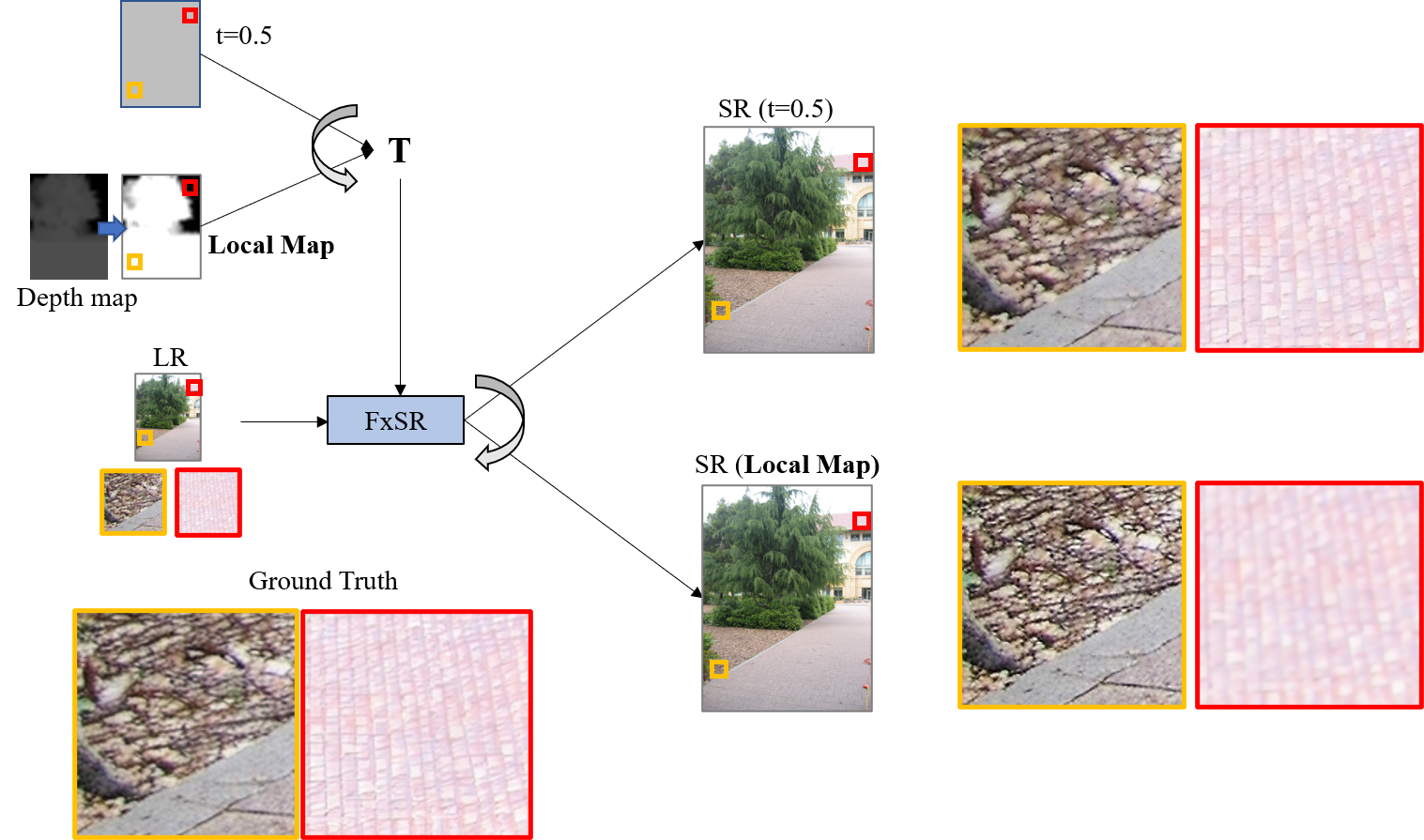}}\vfill
\end{minipage}
\caption{Depth-adaptive FxSR. $\mathbf{T}$-maps is the modified version of the depth map of an image from the Make3D dataset~\cite{saxena2005learning}}.
\label{fig:fig_Local_Map4}
\end{figure*}

%%%%%%%% figure Local Map %%%%%%%% %%%%%%%% %%%%%%%% 
% \begin{figure}[!t]
\begin{figure*}[!t]
\centering
\scriptsize
\begin{minipage}[t]{1.0\linewidth}
    \centering
    {\includegraphics[width=0.85\linewidth
    ]{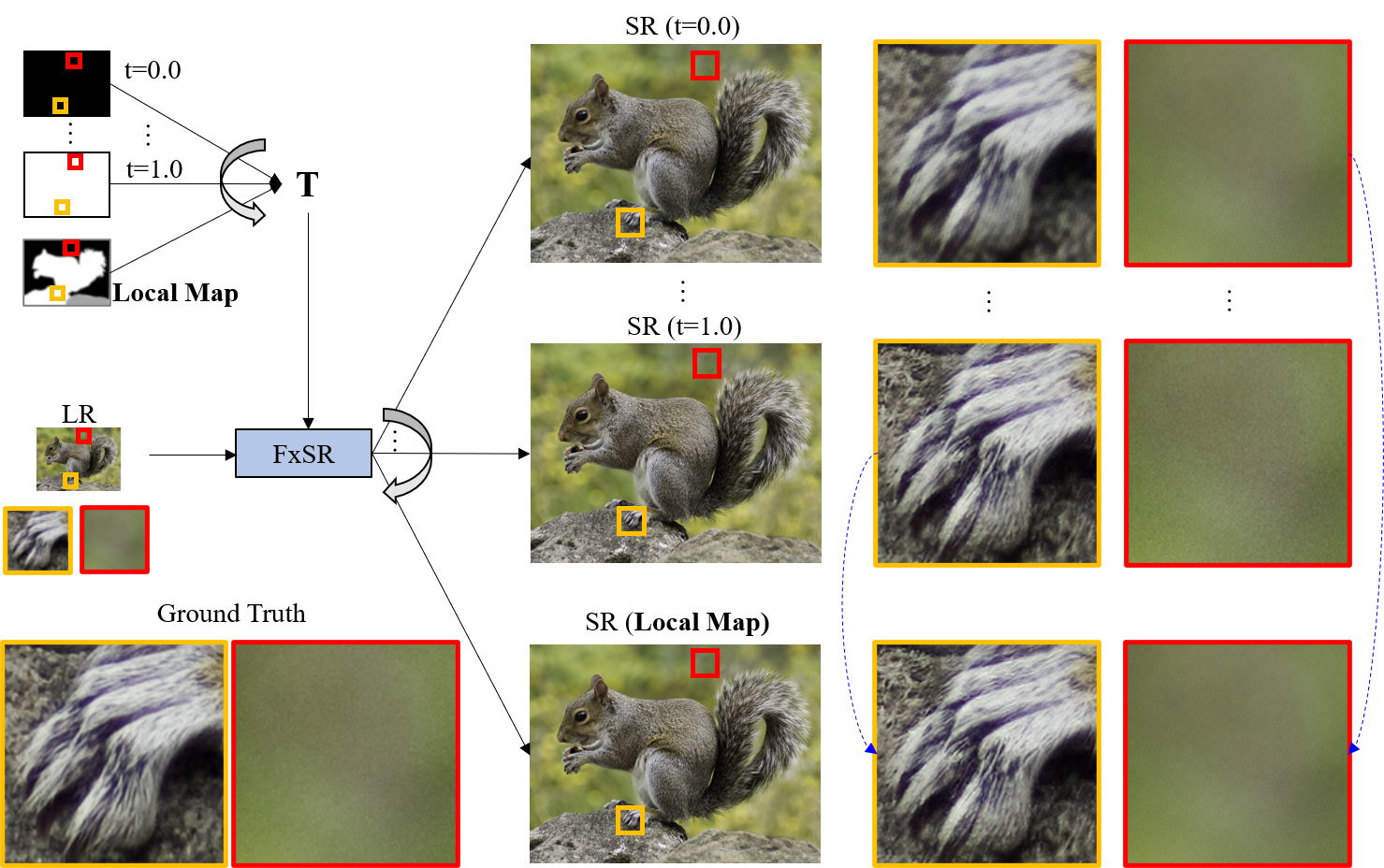}}\vfill
\end{minipage}
\caption{An example of applying a user-created depth map to enhance the perspective feeling with the sharper and richer textured foreground and the background with more reduced camera noise than the ground truth.}
\label{fig:fig_Local_Map3}
\end{figure*}

\subsubsection{Qualitative Comparison} 
Visual comparison between our proposed FxSR-PD and other state-of-the-art methods for 4$\times$ and 8$\times$ are shown in Figures ~\ref{fig:fig_comp_4x} and ~\ref{fig:fig_comp_8x}, respectively. 
We can see that our FxSR-PD provides stronger edges and fine details than the distortion-oriented method RRDB ~\cite{2018esrgan}, and other perception-oriented ones. Also, there are fewer artifacts in our method compared to others.
% %%%%%%%% figure 12 FxSR-CA %%%%%%%% %%%%%%%% %%%%%%%% 
\begin{figure*}[!t]

\setlength{\arrayrulewidth}{1.0pt}
\newcolumntype{Z}
{>{\centering\arraybackslash}X}
\begin{center}
\small
%\footnotesize
\renewcommand{\tabcolsep}{1pt}

\begin{tabularx}{\linewidth}{Z Z Z Z Z Z Z}
\hline
    Whole image  & HR & SRResNet~\cite{2017photo} & SRGAN~\cite{2017photo} & SRGAN-CA & \multicolumn{2}{c}{FxSR-CA}\\
\hline
&&&&&\multicolumn{2}{c}{\includegraphics[width=0.20\linewidth]{figure/control_bar.png}}\\
    &&&&& $t=0.0$ & $t=1.0$ \\
% \hline
\end{tabularx}
\end{center}

\centering
\subfigure[For compressed LR images, FxSR-CA can generate different styles of textures without amplifying artifacts.]{

\begin{minipage}[t]{1.0\linewidth}
    \centering
    \vspace{-0.2cm}
    \includegraphics[width=0.135\linewidth]{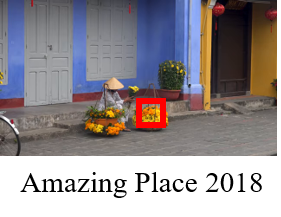}\hfill
    \includegraphics[width=0.135\linewidth]{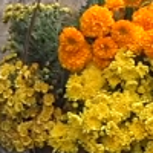}\hfill
    \includegraphics[width=0.135\linewidth]{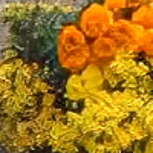}\hfill
    \includegraphics[width=0.135\linewidth]{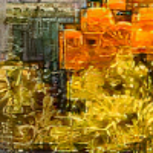}\hfill
    \includegraphics[width=0.135\linewidth]{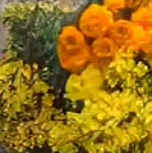}\hfill
    \includegraphics[width=0.135\linewidth]{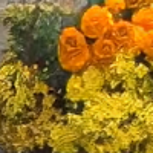}\hfill
    \includegraphics[width=0.135\linewidth]{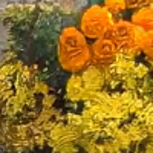}\vfill
    \vspace{0.3cm}
\end{minipage}}

\begin{center}
\renewcommand{\tabcolsep}{1pt}
\begin{tabularx}{\linewidth}{Z Z Z Z Z Z Z}
\hline
    \small Whole image & \multicolumn{3}{c}{\small FxSR-CA} &\multicolumn{3}{c}{\small FxSR-CA}\\
\hline

    &\multicolumn{3}{c}{\includegraphics[width=0.33\linewidth]{figure/control_bar.png}}&\multicolumn{3}{c}{\includegraphics[width=0.33\linewidth]{figure/control_bar.png}}\\
    & \small $t=0.0$ & \small $t=0.5$ & \small $t=1.0$ & \small $t=0.0$ & \small $t=0.5$ & \small $t=1.0$ \\
% \hline
\end{tabularx}
\end{center}

\centering
\subfigure[The intensity and style of textures change according to $t$.]{
\begin{minipage}[t]{1.0\linewidth}

    % \vspace{0.2cm}
    \includegraphics[width=0.135\linewidth]{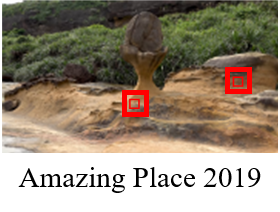}\hfill
    \includegraphics[width=0.135\linewidth]{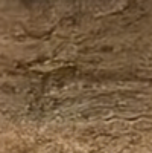}\hfill
    \includegraphics[width=0.135\linewidth]{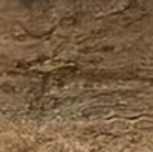}\hfill
    \includegraphics[width=0.135\linewidth]{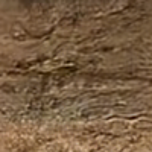}\hfill
    \includegraphics[width=0.135\linewidth]{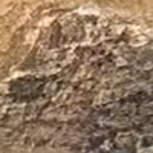}\hfill
    \includegraphics[width=0.135\linewidth]{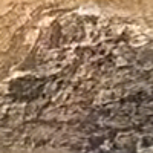}\hfill
    \includegraphics[width=0.135\linewidth]{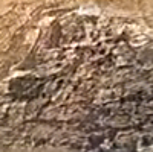}\vfill

\end{minipage}}
\caption{The SR Results for compressed LR images. Two feature space (VGG44 and VGG54) and 16 RBs with SFT are used for FxSR-CA model. LR Images are extracted from "Amazing Place" video title that is encoded by VP9 codec at 0.3Mbps.}
\label{fig:FxSR-CA}
\end{figure*}

\subsection{Flexible SR for Diverse Styles (FxSR-DS)}
\subsubsection{Diverse Style HR Generation}
Unlike the FxSR-PD that attempts flexible trade-offs between perception and distortion, the FxSR-DS aims to generate various styles of HR textures with perceptually high scores for all $t$ values. As shown in Figures from ~\ref{fig:fig_LPIPS_4x} to ~\ref{fig:fig_NIQE_4x}, the FxSR-DS scores better overall with a relatively narrow dynamic range regarding the perception-oriented metrics other than VIF ~\cite{sheikh2006image}. On the other hand, it scores relatively lower for distortion-oriented metrics as in Figure ~\ref{fig:fig_PSNR_4x}. The loss terms and their weights for the conditional objective of the FxSR-DS model are described in Figure ~\ref{fig:FxSR_w}. Different from FxSR-PD with one perceptual loss term, four perceptual loss terms at different feature levels are used. In Figure ~\ref{fig:fig_FxSR-DS_t}, we can see that the SR results for different $t$ values have different types of styles that are clearly distinct from each other. While Figure ~\ref{fig:fig_FxSR-PD_t} shows the trade-off results between perception and distortion, Figure ~\ref{fig:fig_FxSR-DS_t} visualizes our method's scalability to generate various styles of textures by employing more feature spaces into the loss. 

\subsubsection{Quantitative Comparison}
Table ~\ref{tab:tab_div} compares with DNI ~\cite{wang2019deep} and SRFlow ~\cite{2020srflow} in terms of LRPSNR (low-resolution PSNR), LPIPS and Diversity metrics which are evaluation protocol on the Ntire 2021 Challenge ~\cite{SRSpace2021, Lugmayr_2021_CVPR} stated previously. Table ~\ref{tab:tab01} is the evaluation of SR results for a specific $t$ value, while Table ~\ref{tab:tab_div} is the average of all of the SR results for $11$ different $t$ values, from $0$ to $1$, with the step size of $0.1$. Specifically, in Table ~\ref{tab:tab_div}, the FxSR-DS generally scores the best mean LPIPS and Local best (L-best) LPIPS, while the FxSR-PD achieves the best Global best (G-best) LPIPS score. This proves that the perceptually distinct diverse SR results generated by FxSR-DS in Figure ~\ref{fig:fig_FxSR-DS_t} are of high quality in terms of perception-oriented metrics. Since Local Best LPIPS is the maximum performance of the SR model in terms of perceptual measurement, the proposed FxSR-DS shows an improvement of about $2.7\%$ compared to the SRFlow. Figure ~\ref{fig:fig_DS_DIV_LOC} also demonstrates that while the FxDR-PD scores better G-best LPIPS compared to FxDR-DS, the FxDR-DS scores rather superior L-best LPIPS than FxSR-PD. Meanwhile, the SRFlow ~\cite{2020srflow} produces the highest diversity, which learns the sample distribution during training while the proposed models are trained to optimize objectives in the training distribution of objective. However, it is also important to note that the diversity scores are normalized by the G-best as Eqn. ~\ref{eqn:div_score}. This means that the higher the G-best LPIPS, that is, the lower the absolute perceptual quality level, the higher the diversity score.
 
\subsection{Per-pixel Style Control}
In this section, we demonstrate some examples of applying local style control. First, Figure~\ref{fig:fig_Local_Map1} is an example where the LR image has both text and texture areas. In the conventional methods for the SR of Figure~\ref{fig:fig_Local_Map1}(a), multiple SR models are trained with one objective each. Then a model is selected, and the entire image is optimized with the model's objective. If the SR model 0 is selected, which is RRDB~\cite{2018esrgan} representing the distortion-oriented model, the textures of the clothes are blurred while the text edges are restored without artifacts. Conversely, suppose we select the SR model $N-1$, which is ESRGAN~\cite{2018esrgan} representing the perception-oriented model. In that case, some characters in the text area are broken while the textures of the clothes are naturally restored. On the other hand, the proposed FxSR-PD in Figure~\ref{fig:fig_Local_Map1}(b) can restore both the textures of clothes and characters at the same time by applying different objectives to each area through the locally-manipulated style map.

As the second example, let us consider the structural edges of the building and textures of the tree area in Figure ~\ref{fig:fig_Local_Map2}. In a typical approach of using multiple SR models in Figure~\ref{fig:fig_Local_Map2}(a), when the SR model 0 (RRDB) is selected, the structural edges of the building are restored without artifacts, but the tree textures are blurred. Conversely, if the SR model N-1 (ESRGAN) is chosen, the overshoot side-effect occurs around the edges. As shown in Figure~\ref{fig:fig_Local_Map2}(b), similar to the previous example, when a properly adjusted local style map is fed along with the input image, the proposed model FxSR-DS can restore both the tree textures and building edges naturally.

The next is an example of enhancing the perspective feeling when depth information is available, as shown in Figure ~\ref{fig:fig_Local_Map4}. Precisely, input image and depth map pairs used in this example are from the Make3D data set~\cite{saxena2005learning, saxena20083}. When the distance map is used as $\mathbf{T}$ in our FxSR, the foreground region is super-resolved in a perception-oriented way (with emphasized texture), and the background region in distortion-oriented (somewhat blurry). Depth information obtained by some equipment such as Kinect ~\cite{izadi2011kinectfusion} and Time-of-Flight (ToF) camera ~\cite{cui2012algorithms, cui20103d}, or depth estimation algorithms ~\cite{ming2021deep} can be used. It is also possible for users to directly generate a depth map from an input image using image editing S/W, as shown in Figure ~\ref{fig:fig_Local_Map3}. This makes the foreground clearer with sharp details and avoids the unnaturalness of the background becoming as sharp as the foreground. In addition, the camera noise in the background can be reduced. As seen in the examples so far, the proposed method can be used for most cases in various fields that require different processing for each area for a specific purpose.

\subsection{Compressed LR Image Restoration}
Since real-world SR is challenging due to unknown degradation and various noise~\cite{zhang2020ntire, lugmayr2020ntire, nah2019ntire, hussein2020correction, ahn2020simusr, fritsche2019frequency, zhang2018learning, lugmayr2019unsupervised}, we also validate the effectiveness of our method for compressed inputs in Figure~\ref{fig:FxSR-CA}. Unlike previous experiments, FxSR and SRGAN~\cite{2017photo} are re-trained using LR images compressed with JPEG quality factor 90, called FxSR-CA (compression artifacts) and SRGAN-CA. We can see that while compression artifacts are amplified in the results of SRResNet ~\cite{2017photo} and SRGAN ~\cite{2017photo} trained with clean images, the proposed FxSR-CA, generates different style and details according to the change of $t$. To test the effectiveness of the proposed method for the case of real-world compressed images, two videos \footnote{URLs of Amazing Place 2018 and Amazing Place 2019: \url{https://www.youtube.com/watch?v=37IqCYVUhcs}, \url{https://www.youtube.com/watch?v=g5hA2qo2EFc}} which are filmed, edited and copyrighted by Milosh Kitchovitch are used by courtesy of him. Details of the video are provided in the Table~\ref{tab:comp_video}.

%%%%%%%%% TABLE 3 %%%%%%%% %%%%%%%% %%%%%%%% %%%%%%%% %%%%%%%% %%%%%%%%

\begin{table}[ht]
\caption{The details of video we used for SR performance comparison.}
\begin{center}
\small
\begin{tabular}{|c|c|c|}
\hline
Title & Resolution & Bitrate/Codec \\
\hline\hline
Amazing Place 2018    & 640$\times$360	&  319kbps/VP9 \\ 
\hline
Amazing Place 2019    & 640$\times$360	&  301kbps/VP9 \\ 
\hline
% \end{tabularx}
\end{tabular}
\end{center}
\label{tab:comp_video}
\end{table}

% %%%%%%%% Table 3 tab_time_space %%%%%%%% %%%%%%%% %%%%%%%% 
\begin{table}[ht]
\caption{Comparisons of the running time, the computational costs and the size of SR models for super-resolution $4\times$, when the size of LR input images is ${128}\times{128}$.}
\begin{center}
\begin{tabular}{|c||c|c|c|c|}
\hline
{}  & Run Time  & Mult-Add & Param Size &  Forward \\
{}  & (msec)  & $\#$ (G) & (MB) &  Pass (MB)\\
\hline
\hline
{SRGAN~\cite{2017photo}} & 0.014 & 1.51 & 41.63 & 585.11\\
{ESRGAN~\cite{2018esrgan}} & 0.138 & 16.69 & 293.97 & 2061.50\\
% {SRFlow~\cite{2020srflow}} & 0.579 & 39.5 & 158.16 \\
{FxSR} & 0.501 & 18.30 & 320.20 & 8432.78\\
\hline
\end{tabular}
\end{center}
\label{tab:tab_time_space}
\end{table}

\subsection{Complexity Analysis}
We compare the running time, computation costs, and storage size of our methods with other SR methods in Table ~\ref{tab:tab_time_space}. We measure the complexity for the SR $4\times$ processing of one ${128}\times{128}$ LR input image on the environment of NVIDIA RTX3090 GPU. According to Table ~\ref{tab:tab_time_space}, ESRGAN with high-complexity RRDB architecture in Figure ~\ref{fig:f05}(b) requires about 10 times the number of Mult-Add and Run-time than SRGAN. Compared to ESRGAN, FxSR with the proposed RRDBs with SFT in Figure ~\ref{fig:f05}(c) has almost the same number of Mult-Adds and parameter size, but the Forward Pass Size is about 4 times, and the run-time is also increased by 4 times due to the additional memory usage related to the SFT layers. However, it needs to be noted that we use a single network for diverse output generation, whereas the existing methods need at least two networks for producing varying outputs. This is specifically observed in Figure 9, where it is observed that the FxSR requires less or comparable parameters than the network/image interpolation methods that use multiple ESRGAN models.
%-------------------------------------------------------------------------
 %%%%%%%% figure Diversity graph ablation %%%%%%%% %%%%%%%% %%%%%%%% 
% \begin{figure}[!t]
\begin{figure*}[!t]
\centering
\scriptsize
\begin{minipage}[t]{1.0\linewidth}
    \centering
    \hfill
    % \subfigure[] {\includegraphics[width=0.32\linewidth
    % ]{figure/fig08_a_itr_diversity_2.png}}\hfill
    \subfigure[] {\includegraphics[width=0.32\linewidth
    ]{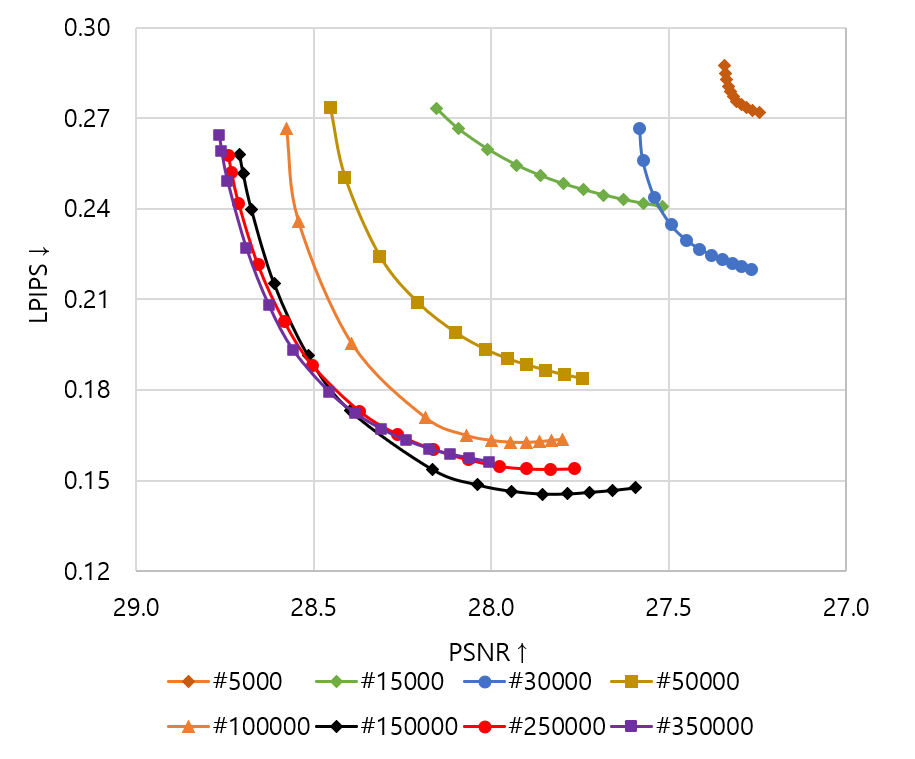}}\hfill
    \subfigure[] {\includegraphics[width=0.32\linewidth
    ]{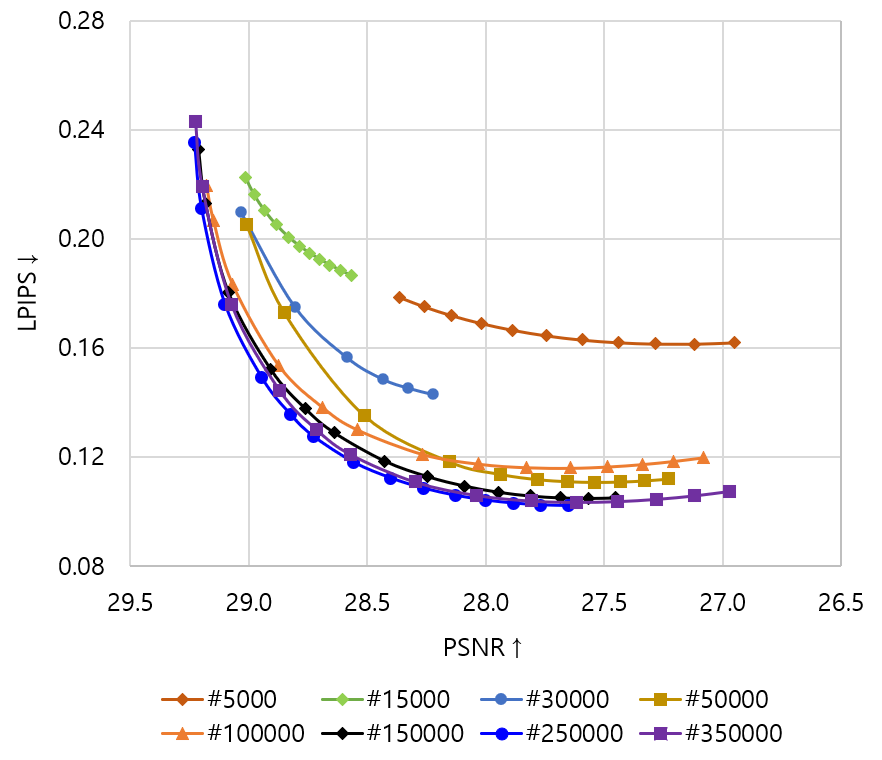}}\hfill
    \subfigure[] {\includegraphics[width=0.32\linewidth
    ]{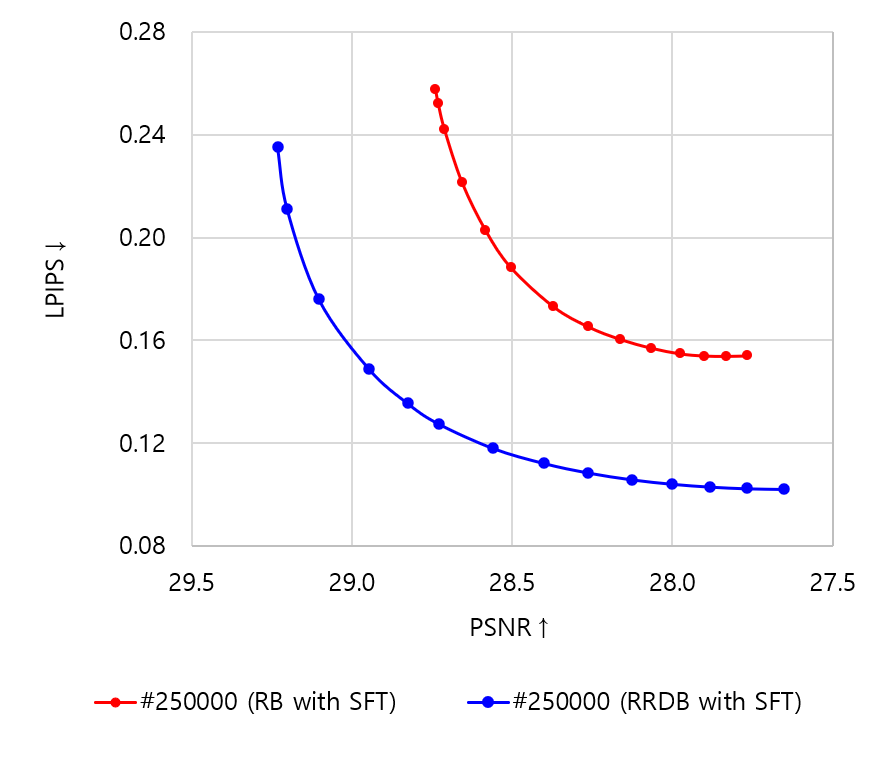}}\hfill
    \hfill
\end{minipage}
\caption{Convergence of diversity curve of the proposed FxSR-PD model as the number of training iteration increase, using (a) 16 RBs with SFT and (b) using 23 RRDBs with SFT. (c) The performance comparison between two FxSR-PD version at the 250,000th iteration}
\label{fig:fig_div_graph_ab}
\end{figure*}
\subsection{Ablation Study}
The goal of classic multi-objective optimization is to find a set of solutions as close as possible to Pareto optimal front and as diverse as possible ~\cite{deb2011multi,lin2019pareto}.
To investigate the performance depending on network architecture and complexity, we observe the change in the perception and distortion (PD) curve while training two versions of FxSR-PD using 16 RBs with SFT in Figure ~\ref{fig:f05}(a), and 23 RRDBs with SFT in Figure ~\ref{fig:f05}(b), respectively. As the number of training iterations increases, the PD curve of FxSR-PD converges to the desired place (lower left), and at the same time, the possible SR range on the curves is also expanded as shown in Figures ~\ref{fig:fig_div_graph_ab}(a) and (b). However, after a certain amount of iterations, the performance does not improve further. Figure~\ref{fig:fig_div_graph_ab}(c) shows the performance comparison between the two FxSR-PD versions at the 250,000th iteration.

\subsection{Discussion}
\subsubsection{Benefits of FxSR}
A single FxSR model can produce different styles corresponding to employed feature losses and is also able to generate intermediate results between the different styles. Moreover, we can control the local regions differently by feeding a control map to the network. Hence, we can have more natural SR outputs by focusing on the foreground or salient regions more than the backgrounds, using user-edited or automatically generated segmentation/depth/saliency maps. Also, we can remedy unnaturally generated regions by controlling the parameters as the post-processing step.

\subsubsection{Limitations of FxSR}
As shown in Table 2, our method can generate comparable or superior results to the existing methods in terms of perceptual quality. But it shows a lower diversity score than the SRFlow because flat control maps are tried in this experiment. Hence, we need more studies on effective control map generation along with other feature spaces and their combinations to increase diversity.

\subsubsection{Future works}
We have used a one-dimensional control parameter $t$ for adjusting SR styles in this work. By defining more than one-dimensional SR style space with various style objectives, we can explore the $n$-dimensional SR spaces, possibly producing more diverse styles. Also, we may consider expanding the work to the image denoising and deblurring to control the degree of restoration locally. Furthermore, leveraging meta-learning would make it possible to improve adaptation to new samples and target objectives.

\section{Conclusion}
We have presented a novel training method and a network structure for the SISR, enabling us to explore various region-wise HR outputs. From this, we can flexibly reconstruct the images between perception-oriented and distortion-oriented ones. This is achieved by defining a conditional objective function with the weights related to the perceptual losses in various feature space levels. Also, our network is designed to modulate the network's intermediate features to change the operation according to these control inputs. As a result, we can generate an image with a desired restoration style for each area. Experiments show that the proposed FxSR yields state-of-the-art perceptual quality and higher PSNR than other perception-oriented methods. Also, we can find many solutions by controlling a single parameter at the inference phase. We will release our code for further research and comparisons.

%Bibliography
\bibliographystyle{IEEEtran}
\bibliography{IEEEabrv, egbib_my}

\end{document}